\documentclass{article}

\usepackage{natbib}
\usepackage{microtype}
\usepackage{graphicx}
\usepackage{booktabs} 
\usepackage{multirow}
\usepackage{amsmath}
\usepackage{amssymb}
\usepackage{mathtools}
\usepackage{amsthm}
\usepackage{amsfonts}
\usepackage{booktabs} 
\usepackage{subcaption}
\usepackage{bbm}
\usepackage[textsize=tiny]{todonotes}
\usepackage{tikz}
\usepackage{pgfplots}
\usepackage{adjustbox}
\usepackage[utf8]{inputenc}
\usepackage[accepted, nohyperref]{icml2022}

\theoremstyle{plain}

\theoremstyle{definition}

\theoremstyle{remark}

\usetikzlibrary{bayesnet}
\usepgfplotslibrary{units}
\usepgfplotslibrary{groupplots}
\pgfplotsset{grid style=none}
\pgfplotsset{axis y line=left}
\pgfplotsset{axis x line=bottom}

\newcommand{\GP}[0]{\mathcal{GP}}
\newcommand{\E}[0]{{  \mathbb{E}   }}
\newcommand{\cov}[0]{\text{cov}}

\newcommand{\x}[0]{\mathbf{x}}
\newcommand{\X}[0]{\mathbf{X}}
\newcommand{\tk}[0]{\tilde{k}}
\newcommand{\w}[0]{\mathbf{w}}

\newcommand\HUGE{\@setfontsize\Huge{50}{65}} 
\newlength\figurewidth
\newlength\figureheight
\RequirePackage{luatex85,shellesc}
\icmltitlerunning{Additive GPs Revisited}

 \usepackage[hyphens]{url}
 \usepackage[pagebackref=true]{hyperref}
\renewcommand*\backref[1]{\ifx#1\relax \else (Cited on #1) \fi}
  \hypersetup{
    breaklinks=true,   
    colorlinks=true,   
    pdfusetitle=true,  
 }
 \usepackage[hyphenbreaks]{breakurl}

\begin{document}

\twocolumn[
\icmltitle{Additive Gaussian Processes Revisited}
\icmlsetsymbol{equal}{*}

\begin{icmlauthorlist}
\icmlauthor{Xiaoyu Lu}{comp}
\icmlauthor{Alexis Boukouvalas}{comp}
\icmlauthor{James Hensman}{comp}
\end{icmlauthorlist}

\icmlaffiliation{comp}{Amazon, Cambridge, United Kingdom}

\icmlcorrespondingauthor{Xiaoyu Lu}{luxiaoyu@amazon.com}
\icmlkeywords{Machine Learning, ICML}

\vskip 0.3in
]
\printAffiliationsAndNotice{}

\begin{abstract} 
Gaussian Process (GP) models are
a class of flexible non-parametric models that have rich representational power. By using a Gaussian process with additive
structure, complex responses can be modelled whilst retaining interpretability. Previous work showed that additive Gaussian process models
require high-dimensional interaction terms. We propose the orthogonal
additive kernel (OAK), which imposes an orthogonality constraint on the
additive functions, enabling an identifiable, low-dimensional representation of the functional relationship. We connect the OAK kernel to functional ANOVA decomposition, and show improved convergence rates for sparse computation methods. 
With only a small number of additive low-dimensional terms, we demonstrate the OAK model achieves similar or better predictive performance compared to black-box models, while retaining interpretability.
\end{abstract}

\section{Introduction}

\begin{figure*}[h!]
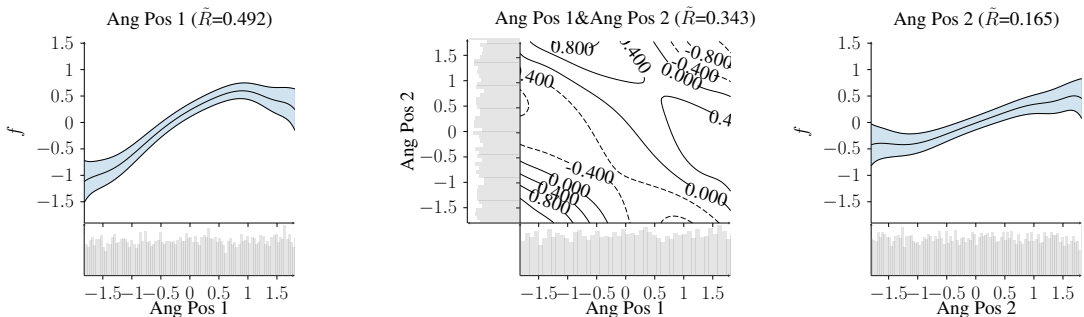

	\vspace*{-0.1in} 
	\centering
	\begin{subfigure}[b]{0.3\textwidth}
		\input{tikz/UCI/pumadyn/pumadyn_feature1.tex}
	\end{subfigure} 
	\begin{subfigure}[b]{0.3\textwidth}
		\input{tikz/UCI/pumadyn/pumadyn_feature12.tex}
	\end{subfigure} 
	\begin{subfigure}[b]{0.3\textwidth}
		\input{tikz/UCI/pumadyn/pumadyn_feature2.tex}
	\end{subfigure} 
	\vspace*{-0.2in} 
	\caption{Visualization of the decomposed functions with highest Sobol indices for the pumadyn dataset. 
    On the horizontal axis we plot different feature $x_i$, on the vertical aixs is its corresponding function $f_i$. 
    For two-way interaction terms, a contour plot is used. Grey bars represent histograms of input features, 
    black solid lines represent posterior mean GP, blue shaded area represents $\pm$ 2 standard deviations 
    confidence interval from the GP model. $\tilde{R}$ in the brackets represent (normalized) Sobol indices. 
    We can observe that over $99\%$ of the variance can be explained with only these three terms: 
    two first order terms and one interaction term between them. We reach optimal model performance with only these three terms (Figure \ref{fig:sobol_plots}).
    }
	\label{fig:pumadyn}
\end{figure*}

Gaussian Processes (GPs) can be used to construct additive models by using the
property that a sum of two GPs results in a new GP with a kernel function
defined as the sum of the original ones. Using an additive structure in a
Gaussian process model is enticing from an explainability standpoint, since one
can use the linear properties of the GP to perform inference over the added
components, which can yield insights into the data.  
For datasets with more than one input dimension, it is straight-forward to build GP models as a sum of one-dimensional functions, or known pairs (or triplets, etc.)
of interacting inputs. In the statistics literature, Generalized Additive Models (GAMs)
\citep{hastie2017generalized, wood2017generalized}, are often built using sums of splines over either each input independently or over carefully selected sets of inputs. 

From the standpoint of explainable and interpretable machine learning, additive
Gaussian processes such as those considered within \citet{kaufman2010bayesian, duvenaud2011additive, timonen2019lgpr} offer the promise of automatically discovering relevant
features and combinations of features, through learning of a kernel with
parameterized additive structure. In particular, \citet{duvenaud2011additive} proposed a kernel which allows additive interactions of all orders, ranging from first order terms to the interactions between all the features. An efficient computation scheme was proposed for avoiding the exponentially large sum required
over combinations. 

In this work, we build on and challenge the findings of
\citet{duvenaud2011additive}, where the experimental results suggest
that high order terms are required to model some of the regression and classification
datasets. We show that the dimensionality of the models constructed
is considerably higher than necessary: for example, their model of the 8-dimensional  {\it pumadyn} dataset requires an 8-dimensional interaction 
whereas our proposed model requires only 2-dimensional interactions (see Figure \ref{fig:pumadyn}). 
A full comparison on all the datasets used in \citet{duvenaud2011additive} is provided in Section
\ref{sec.duvenaud_experiments}: in all cases, we find that a small number of
low-dimensional terms are needed to achieve similar or better performance. We posit that the high dimensional nature of their models are due
to two issues: an identifiability issue with the summed components; and the way
that the contribution of a component to the overall model is measured. 

We solve the identifiability issue by borrowing an idea from
\citet{durrande2011additive}, where the components of the additive model are
orthogonalized. We call the resulting kernel \emph{orthogonal additive kernel}
(OAK),  which can produce highly parsimonious models of the datasets studied in
\citet{duvenaud2011additive}, as well as more recent larger datasets.
\citet{plumlee2018orthogonal} tackles a slightly different identifibality issue by proposing a GP whose
stochastic part is orthogonal to the mean part. We measure
the contribution of any component to the overall model using a Sobol index
\citep{sobol1993sensitivity, owen2014sobol} which is shown to be analytic for the OAK model. We see in Section
\ref{sec.duvenaud_experiments} that the pumadyn dataset can be modelled using a
sum of only three components -- one two-dimensional interaction function and two
one-dimensional functions. These parsimonious models are highly explainable
since each effect of a component can be examined in a simple plot, yet the model remains powerful: the predictive performance is on par with or better than
either the original additive model or a full squared exponential GP model. In a
case study on the SUSY physics dataset (Section \ref{sec.susy}), our method
produces a low dimensional model with only ten one-dimensional and
two-dimensional terms that outperforms the dropout-based neural network
baseline. On another case study of a contemporary dataset of customer churn
(Section \ref{sec.churn}), our method outperforms the XGBoost baseline whilst
providing low-dimensional components that offer insights into
business problems. 

Finally, since the OAK method is using a new kernel within a standard GP
formulation, we are able to scale the method using sparse GP
methods. We show in Section \ref{sec.inducing} that the scalability of a sparse
GP with the OAK kernel is favorable to that of a squared
exponential kernel, since the eigenspectrum of our low dimensional model is more
easily represented by an inducing point formulation. We build on recent work
\citep{burt2019rates} to show increased convergence rates for sparse GPs with
our proposed kernel. 

Our main contribution is to combine the orthogonality constraint in \citet{durrande2011additive} with the additive model in \citet{duvenaud2011additive} that utilizes the Newton-Girard trick, where computationally complexity scales polynomially rather than exponentially with the number of features. We draw the link to functional ANOVA (FANOVA) decomposition \cite{owen2014sobol, chastaing2015anova, ginsbourger2016anova}  and quantify the contribution of each component with analytic Sobol indices. We have conducted extensive sets of regression and classification experiments to show its practical value. The resulting model is parsimonious and interpretable, requiring minimal model tuning.

The remainder of this manuscript is organized as follows. In Section
\ref{sec:duvenaud} we recap the additive model used in
\citet{duvenaud2011additive} and propose OAK in Section \ref{sec:orthogonality}.
We introduce Sobol index in Section \ref{sec:sobol} and discuss its relationship
with functional ANOVA decomposition and OAK. Experimental results are given in Section
\ref{sec:experiment} and we conclude in Section \ref{sec:conclusion}.
Our code is available at \url{https://github.com/amzn/orthogonal-additive-gaussian-processes}.

\section{Background and Related Work}
\label{sec:duvenaud}
We are interested in modeling output $y$ as a function of $D$-dimensional input features
$\x:=(x_1, \cdots x_D)$ with a hidden function $f(\x)$.
\citet{duvenaud2011additive} considers building a GP model with the additive structure:
\begin{align}
f(\x) &= f_1(x_1) + f_2(x_2) + \cdots + f_{12}(x_1,x_2) \nonumber \\
&+ \cdots + f_{12...D}(x_1,x_2,\cdots x_D).
\label{eq:duvenaud}
\end{align}
In a GP model, the additive structure of the function decomposition is enforced
through the structure of the kernel, whose decomposition can be constructed as
follows: first assign each dimension $i \in \{1 . . . D\}$ a one-dimensional
\emph{base kernel} $k_i(x_i, x_i')$; then define the first order, second order
and
$d^{th}$ order additive kernel as:
\begin{align}
k_{add_1}(x,x') &= \sigma_1^2\sum_{i=1}^D k_i(x_i, x_i')\,, \nonumber\\
k_{add_2}(x,x') &= \sigma_2^2\sum_{i=1}^D \sum_{j={i+1}}^D k_i(x_i, x_i') k_j(x_j, x_j'), \label{eq:duvenaud_kernel}\\
k_{add_d}(x,x') &= \sigma_d^2\sum_{1\leq i_1 \leq i_2 \leq \cdots \leq i_d \leq D} \left[\prod_{l=1}^d k_{i_l} (x_{i_l}, x_{i_l}') \right],\, \nonumber
\end{align}
with the kernel then constructed by summing over all of the orders up to the dimensionality of the data. The parameters
$\sigma_d^2$ control the relative importance of high-dimensional and
low-dimensional functions in the sum: we shall see later in this work that the
high-order terms can be set to zero for all the datasets we consider using our proposed method,
effectively truncating the sum. Although there can be a very large number of
terms in the kernel, \citet{duvenaud2011additive} proposed an algorithm based on
the Newton-Girard identity to efficiently compute the kernel in polynomial time, see detailed algorithm in Appendix \ref{sec:newton_girard_alg}.

When it comes to measuring the importance of each interaction, \citet{duvenaud2011additive} proposed considering the estimated parameters $\sigma_d^2$. In Section
\ref{sec:orthogonality} we show through a simple example that these parameters
are unidentifiable.  We follow a different approach using 
Sobol indices \citep[e.g.][]{sobol1993sensitivity, muehlenstaedt2012data, owen2014sobol} to weigh the importance of
different components of the construction.  

Imposing an orthogonal constraint on additive kernel components was proposed by
\citet{durrande2011additive} and extended in \citet{durrande2013anova} and \citet{martens2019enabling}. Denoting the constrained kernel by $\tilde{k}$, the kernel was constructed in the form
$k(\mathbf x, \mathbf x') = \prod_{d=1}^D (1 + \tilde k_d(x_d, x'_d))$, which
does not allow for control of the importance of different orders, {\it
cf.} (\ref{eq:duvenaud_kernel}), and they did not apply the kernel in the context
of GP regression, so were not able to learn kernel parameters. \citet{martens2019decomposing} also extended \citet{durrande2011additive}, building low-dimensional latent variable models where the latent and observed features are orthogonal. In the current paper, we focus on the interpretability and parsimony of the orthogonal models for regression and classification tasks in a practical setting. In particular, we extend to large numbers of features through the efficient Newton-Girard procedure of \citet{duvenaud2011additive}.

\section{Orthogonality}
\label{sec:orthogonality}
\begin{figure*}[h!]
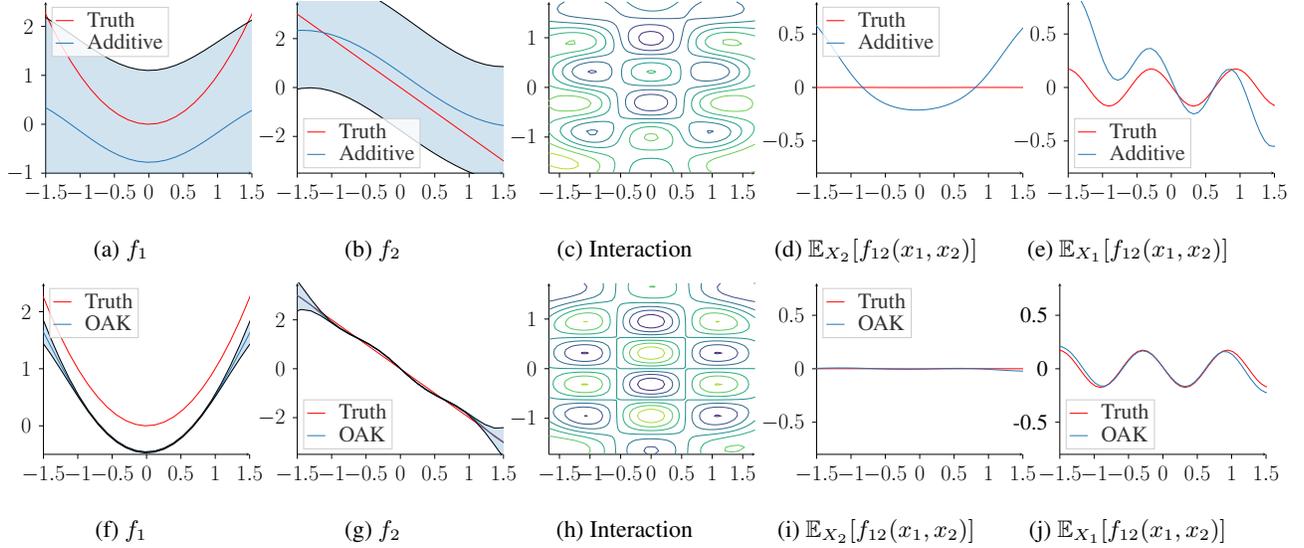

	\captionsetup[subfigure]{aboveskip=-1pt,belowskip=-1pt}
	\centering
	\begin{subfigure}[b]{0.19\textwidth}
		\input{tikz/2d/2d_rbf_f1_0.tex}
		\caption{$f_1$}
		\label{fig:rbd_f1}
	\end{subfigure} 
	\begin{subfigure}[b]{0.19\textwidth}
		\input{tikz/2d/2d_rbf_f2_0.tex}
		\caption{$f_2$}
		\label{fig:rbd_f2}
	\end{subfigure}
	\begin{subfigure}[b]{0.19\textwidth}
		\input{tikz/2d/2d_interaction_rbf_0.tex}
		\caption{Interaction}
	\end{subfigure}	
	\begin{subfigure}[b]{0.19\textwidth}
		\input{tikz/2d/2d_rbf_interaction_f1_0.tex}
		\caption{$\E_{X_2}[f_{12}(x_1,x_2)] $}
		\label{fig:inter_rbf_f1}	
	\end{subfigure}	
	\begin{subfigure}[b]{0.19\textwidth}
		\input{tikz/2d/2d_rbf_interaction_f2_0.tex}
		\caption{$\E_{X_1}[f_{12}(x_1,x_2)]$ }
		\label{fig:inter_rbf_f2}
	\end{subfigure}

\vspace{0.1in}
	\begin{subfigure}[b]{0.19\textwidth}
		\input{tikz/2d/2d_f1.tex}
		\caption{$f_1$}
		\label{fig:oak_f1}
	\end{subfigure} 
	\begin{subfigure}[b]{0.19\textwidth}
		\input{tikz/2d/2d_f2.tex}
		\caption{$f_2$}
	\end{subfigure}
	\begin{subfigure}[b]{0.19\textwidth}
		\input{tikz/2d/2d_interaction_Durrande.tex}
		\caption{Interaction}
	\end{subfigure}	
	\begin{subfigure}[b]{0.19\textwidth}
		\input{tikz/2d/2d_interaction_f1.tex}
		\caption{$\E_{X_2}[f_{12}(x_1,x_2)] $}
		\label{fig:inter_f1}
	\end{subfigure}	
	\begin{subfigure}[b]{0.19\textwidth}
		\input{tikz/2d/2d_interaction_f2.tex}
		\caption{$\E_{X_1}[f_{12}(x_1,x_2)]$ }
		\label{fig:inter_f2}
	\end{subfigure}
	\caption{Illustration of the non-identifiability of the additive GP model in \citet{duvenaud2011additive} on the two-dimensional problem. Top row: additive GP model; bottom row: OAK model.  
	Red and blue lines represent the true and learned posterior mean functions respectively, blue shaded area represent $\pm2$  standard deviation. From left to right:	posterior of $f_1$ and $f_2$; posterior mean of $f_{12}$;
	marginal plot for $f_1$ in the interaction term ($\E_{X_2}
	[f_{12}(x_1,x_2)]$); marginal plot for $f_2$ in the interaction term
	($\E_{X_1} [f_{12}(x_1,x_2)]$). Note how the quadratic shape in Figure
	\ref{fig:rbd_f1} and the linear trend in Figure \ref{fig:rbd_f2} are
	captured in the higher order terms (Figure \ref{fig:inter_rbf_f1}  and
	\ref{fig:inter_rbf_f2}). OAK correctly identifies the true additive components with smaller uncertainties. Note that the constant gap between the truth and OAK in Figure \ref{fig:oak_f1} is expected and is captured with the constant kernel.}
	\label{fig:2d}
\end{figure*}

With the decomposition in (\ref{eq:duvenaud}), we may learn different models
that give the same predictions: this is due to the non-identifiability of the
summed functions \citep{ginsbourger2008discrete, martens2019enabling}. Assume a two-dimensional problem:
\begin{align} f(x_1,x_2) = f_1(x_1) + f_2(x_2)\,,
\end{align}
with the true functional decomposition $f_1$ and $f_2$, then
\begin{align} f(x_1,x_2) = (f_1(x_1) + \Delta) + (f_2(x_2)  - \Delta)
\end{align}
is a valid decomposition for any value of $\Delta$. In other
words, there are infinitely many possible decompositions of $f$. This is not
desirable because it makes interpretability difficult: which of the
decompositions should one choose?  Moreover, higher order terms can absorb
effects from lower order terms and one may learn a model that is more
complicated than needed, as we will now illustrate. 

Take a two-dimensional example with true decomposition:
\begin{align}
f(x_1,x_2) = x_1^2 - 2x_2 + \cos(3x_1)\sin(5x_2)\,.
\label{eq:2d}
\end{align}
We sample $x_1$ and $x_2$ uniformly on $(-1,1)$ and generate $y \sim f(x_1,x_2)
+ \epsilon$ with $f$ in (\ref{eq:2d}) and $\epsilon \sim \mathcal{N}(0,0.01)$.
We then fit an additive GP model \cite{duvenaud2011additive} with
squared exponential base kernels. We learn the kernel parameters and likelihood
(noise) variance using maximum likelihood.  The experiment is
repeated with 9 random seeds and three unique local optima
(i.e., 3 sets of hyperparameters) are discovered. We show posterior
functions for one of the local optima in Figure \ref{fig:2d} (top) (details in Appendix \ref{sec:toy}).

In Figure \ref{fig:2d} (top) we observe that the functions $f_1$ and $f_2$ have
large (marginal) variance, meaning the model is less certain in isolating
individual effects from other terms. In Figure \ref{fig:inter_rbf_f1}, we plot
the interaction term with respect to $x_1$ by taking the average of
$f_{12}(x_1,x_2)$ over $x_2$, i.e., $\E_{X_2} [f_{12}(x_1,x_2)]$.  
Figure  \ref{fig:inter_rbf_f2} is a similar plot of $x_2$ by marginalizing out $x_1$. The
quadratic shape in Figure \ref{fig:inter_rbf_f1} and the linear trend in Figure
\ref{fig:inter_rbf_f2} show that the interaction term is capturing the
individual effect of $f_1$ and $f_2$. In other words, higher order terms absorb
the effect of lower order terms. The reverse can also be true, see Appendix \ref{sec:toy}.

\subsection{GP with Orthogonal Additive Kernel}
To mitigate the identifiability problem, we incorporate an idea from
\citet{durrande2011additive}, where a constraint is used on each base kernel such
that the integral of each function $\{f_i\}_{i=1}^D$  with respect to the input
measure is zero.  For $f$ with non-zero mean,
the offset can be modelled using a constant kernel, resulting in a unique
decomposition. Our model takes the same form as (\ref{eq:duvenaud}), 
except adding an additional GP $f_0$ with constant kernel. Define $[D]:= \{1,\cdots,D\}$, we constrain each $f_i$ to satisfy the \emph{orthogonality constraint}:
\begin{align}
\int_{\mathcal{X}_i} f_i(x_i)p_i(x_i)dx_i = 0,
\label{eq:orthogonal_constraint}
\end{align}
for $i \in [D]$, where $\mathcal{X}_i$  and $p_i$ are the sample space and the density for input feature $x_i$ respectively.

We now describe how we can construct the kernel for each $f_i$. 
For each feature $i$ with base kernel $k_i$, it can be shown that conditioning
on  $S_i:= \int f_i(x_i)p_i(x_i)dx_i = 0$, the process $f$ is another GP with a
modified kernel $\tilde{k}_i$:
 \begin{align}
f_i(\cdot) \, \bigg | \int f_i(x_i)p_i(x_i)dx_i = 0 \sim \GP(0,\tilde{k}_i),
\end{align}
where
\begin{align}
\tilde{k}_i(x_i,x_i') =&k_i(x_i,x_i') - \E[S_i f_i(x_i)] \E[S_i^2]^{-1} \E[S_i f_i(x_i')]\,, \nonumber\\
\E[S_i f_i(\cdot)] =& \int p_i(x_i) k_i(x_i, \cdot) dx_i\ ,  \nonumber\\
\E[S_i^2] =& \int\int p_i(x_i) p_i(x_i') k_i(x_i,x_i') dx_i dx_i'\,.
\label{eq:durrande_kernel}
\end{align} 
We call $\tilde{k}_i$ the \emph{constrained} kernel. 
For higher order interaction terms, we desire the constraint
$\int_{\mathcal{X}_i} f_u(\x_u)p_i(x_i)dx_i = 0$ $\forall i \in u$ where $\x_u :=
\{x_i\}_{i \in u}$.  This is achieved by simply taking the product of one-dimensional constrained kernels: for any $u \subseteq  [D]$,
\begin{align} 
\tilde{k}_u(x,x') = \prod_{i \in u} \tilde{k}_i(x_i,x_i').
\label{eq:product_kernel}
\end{align}
A function $f_u$ drawn from a GP with the constrained kernel $\tilde{k}_u$ satisfies the
orthogonality condition assuming independent input features, see proof in Appendix \ref{sec:orthogonality_product}.

Since the orthogonal construction can be achieved by using sums
and products of kernels, we can construct our model by plugging in the
constrained kernel \eqref{eq:durrande_kernel} to the sum structure \eqref{eq:duvenaud_kernel}. We call this the \emph{Orthogonal Additive Kernel} (OAK). Note that under the orthogonality constraint, the decomposition in  \eqref{eq:duvenaud_kernel} is identifiable since it is precisely the FANOVA decomposition, see details in Section \ref{sec:sobol}.

\subsection{Base Kernel}
\label{sec:base_kernel}
We choose to use a squared exponential kernel for continuous features as the base kernel due to its analytic solution with orthogonality constraints. Other kernel choices such as the Mat\'{e}rn kernel also leads to analytic expressions for the constrained kernel.

Specifically, for squared exponential base kernel $k_i$ with unit variance and lengthscale $l_i$: $k_i(x_i,x_i') = \exp\left(-\frac{(x_i-x_i')^2}{2l_i^2}\right)$, $\tilde{k}_i$ is analytic and has a closed form solution when the input density $p_i$ is Gaussian, mixture of Gaussian, uniform, categorical, or approximated with the empirical distribution. We hereby give results in the case of Gaussian measure: without loss of generality, assuming one-dimensional $x$ with $p(x) = \mathcal{N}(\mu,\delta^2)$ where we drop subscript $i$ for simplicity, the constrained squared exponential $\tilde{k}$ is:
\begin{align}
\tilde{k}(x,x') := &\exp\left(-\frac{(x-x')^2}{2l^2}\right) - \frac{ l \sqrt{l^2 + 2\delta^2 }}{l^2 + \delta^2} \times  \nonumber\\
& \exp\left(-\frac{((x-\mu)^2 + (x'-\mu)^2)}{2(l^2 + \delta^2)}\right).
\label{eq:crbf}
\end{align}
For other forms of input densities, please refer to 
Appendix \ref{sec:input_measure}. For categorical features, we can use the categorical kernel and an empirical input density $p$ (see Appendix \ref{sec:coregional}  and \ref{sec:empirical_measure}).

\subsection{Normalizing Flow} To satisfy the Gaussian input density assumption,
we use a normalizing flow \citep{rezende2015variational} to transform continuous
input features to have an approximate Gaussian density. This is achieved by
applying a sequence of bijective transformations on each feature, whose
parameters are learnt by minimizing the KL divergence between a standard
Gaussian distribution and the transformed input data. The parameters are
then fixed \emph{before} fitting the OAK model on the transformed data with approximate Gaussian densities. For details and ablation studies, see Appendix \ref{sec:normalising_flow} and \ref{sec:nf}.

\subsection{Illustration}
We use the example from (\ref{eq:2d}) to illustrate the constrained model described above with results given in Figure \ref{fig:2d} (bottom). We have found that the GP model with the constrained squared exponential kernel is more stable as all 9 runs using different initial configurations converge to the same hyperparameters as opposed to the unconstrained model where we have found 3 local optima. We are able to capture the correct form of first order terms and the interaction component, resulting in a better fit and better calibrated uncertainty. Note that the constant gap (vertical shift) in Figure \ref{fig:oak_f1} is expected since we constrained each function to have zero mean with respect to the input density, and a separate constant kernel is used to capture the gap due to the non-zero mean of $f$.

\subsection{Sparse GP with Inducing Points} \label{sec.inducing}
When the number of data points $N$ is big, GP inference costs $\mathcal{O}(N^3)$ in computation which is expensive. Variational inference with sparse GP can be used to reduce the computational costs to  $\mathcal{O}(NM^2)$ where $M$ is the number of inducing variables \citep{titsias2009variational}.

\citet{burt2019rates} showed that the number of inducing points $M$ needed for sparse GP regression with normally distributed inputs in D-dimensional space with the squared exponential kernel is $M = \mathcal{O}(log^D N) $.

\begin{figure}[h!]
	\centering
	\begin{subfigure}{0.45\textwidth}
		\centering
		\hspace{-10mm}
		\input{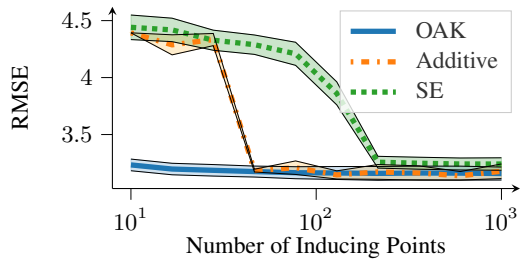}
	\end{subfigure} 
		\vspace*{-0.07in}
	\caption{Test RMSE versus number of inducing points for the pumadyn dataset. Results are averaged over 5 repetitions, shaded area represents $\pm 1$ standard deviation.}
	\label{fig:pumadyn_induing}
\end{figure}

In practice, one can limit the maximum order of interactions to be $\tilde{D} \leq D$. For our additive model, the number of kernels to be added is therefore $\sum_{k=1}^{\tilde{D}}  {D \choose k} $
and the number of inducing points needed is
\[
\sum_{k=1}^{\tilde{D} } {D \choose k}  \mathcal{O}(log^k N) = \mathcal{O}\left({D \choose \tilde{D}} log^{\tilde{D}} N \right).
\]
The number of inducing points needed for OAK is smaller than that for the non-orthogonal case. We also verify this empirically on the pumadyn dataset with a 4:1 training-test split. We compare our model with its non-orthogonal counterpart as in (\ref{eq:duvenaud}) and a sparse GP model with squared exponential kernel. Results are displayed in Figure \ref{fig:pumadyn_induing}, where OAK converges much faster and needs a smaller number of inducing points to reach same/better test RMSE (additional experiments can be found in Appendix \ref{sec:pumadyn_rmse}). 

\section{ANOVA Decomposition and Sobol Indices}
\label{sec:sobol}
Practitioners  are often interested in the importance of features in predicting the output. For example, $f$ may be explained using only a small number of features or interactions despite there being a large number of features. Global sensitivity analysis \citep{saltelli2008global} is a measure of importance of input features, based on an analysis of variance (ANOVA) decomposition. Sobol indices \citep{sobol1990sensitivity} are one such measure for attributing value of an output to individual features. We will see later that the Sobol indices are analytic for the OAK model.

Functional ANOVA (FANOVA) \citep{hoeffding1948central, stone1994use, huang1998projection} decomposes a function $f(\x)$ into the form 
$f(\x) = \sum_{u \subseteq [D]} f_u(\x_u)$,
where $f_u$ only depends on $x_j$ for $j \in u$ and is defined recursively by 
\begin{align}
f_u(\x) = \int_{\mathcal{X}_{-u}} \left(f(\x) - \sum_{v \subset u} f_v(\x_v)\right) dP(\x_{-u}),
\end{align}
where $f_{\emptyset}(\x) = \E[f(\x)]$, $\x_{-u}$ denotes $\x$ excluding $x_u$ and $P(\x)$ denotes the distribution of $\x$. Applying the FANOVA decomposition to our OAK construction in (\ref{eq:duvenaud}) and  (\ref{eq:durrande_kernel}) reveals that the functions considered in OAK are precisely the components of the FANOVA decomposition, see proof in Appendix \ref{sec:anova_proof}. The FANOVA decomposition associates each component with a variance. This variance is due to disturbances on the input to the function: 
we denote it by $\mathbb{V}_x[f_u]$. The orthogonality of OAK leads to the ANOVA identity \citep{owen2014sobol}:
\begin{align}
R := \mathbb{V}_x[f(\x)] = \sum_{u\subseteq[D]} R_u,
\end{align}
where $R_u := \mathbb{V}_x[f_u(\x)]$ is defined as the Sobol index for feature set $u$.  In other words, each $R_u$ measures how much variance is explained by feature set $u$, measuring the importance of the features. We normalize the Sobol indices such that they sum up to 1 and denote the normalized Sobol indices with $\tilde{R}$ in later sections. 
Similarly to \citet{durrande2011additive}, to assess the relative importance of a component of our model, we consider the Sobol index of the posterior mean function associated with that component:
$\tilde{R}_u = \frac{\mathbb{V}_\x [ m_u(\x) ]} { \mathbb{V}_\x [ m(\x)] }$,
where $m_u$ and $m$ denote the posterior mean function of $f_u$ and $f$ respectively. In particular,
\begin{equation}
m_u(x) = \sigma_{|u|}^2\left( \odot_{i \in u} \tilde{k}_i(x_i, \X_i) \right) K(\X,\X)^{-1} \mathbf{y} \label{eq:posterior_component}
\end{equation}
where $K(\X,\X)$ denotes the training input covariance across all inputs, $\X_i$ and $y$ denote the $i$-th column of $\X$ and the 
vector of output observations, $\sigma_{|u|}^2$ is the associated variance parameter for the $|u|$-th order interaction and $ \odot$ denotes element-wise multiplication. A similar formula for sparse GP can also be obtained. 
The Sobol index associated with the input set $u$ is therefore
\begin{align}
&\mathbb{V}_x [ m_u(x) ] = \sigma_{|u|}^4 y^\top K(\X,\X)^{-1}   \odot_{i \in u}  \label{eq:sobol}\\
&\left( \int \tilde{k}_i(x_i, \X_i) \tilde{k}_i(x_i, \X_i)^\top \textrm{d}p_i(x_i) \right) K(\X,\X)^{-1}  \mathbf{y},\nonumber
\end{align}
since $\E_x[m_u(x)] = 0$ due to the orthogonality constraint. In case of 1) constrained squared exponential kernel and a Gaussian measure or 2) binary/categorical kernel with discrete measure, the integral is tractable and can be computed analytically. More details can be found in Appendix \ref{sec:sobol_appendix}. Note that the Sobol index is not affected by our normalizing-flow transformation of the input, see details in Appendix \ref{sec:invariance_sobol}.

\section{Experiments}
\label{sec:experiment}
\addtocounter{footnote}{1} 
\begin{figure*}[t!]
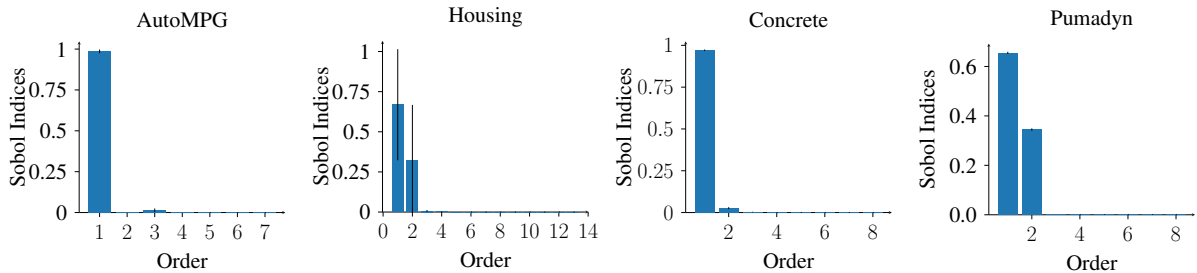

   \vspace*{-6mm}
	\centering
	\centering
	\begin{subfigure}{0.23\textwidth}
		\input{tikz/UCI/autoMPG_sobol_order.tex}
	\end{subfigure} 
	\begin{subfigure}{0.23\textwidth}
		\vspace{3.5mm}
		\input{tikz/UCI/Housing_sobol_order.tex}
	\end{subfigure} 
	\begin{subfigure}{0.23\textwidth}
		\input{tikz/UCI/concrete_sobol_order.tex}
	\end{subfigure} 
	\begin{subfigure}{0.23\textwidth}
		\input{tikz/UCI/pumadyn_sobol_order.tex}
	\end{subfigure} 
	\vspace*{-0.25in}
	\caption[Caption for LOF]{Sum of normalized Sobol indices for each interaction order for UCI regression problems, where bars represent one standard deviation across 5 cross-validation splits. All of the datasets require $\leq 3$ order of interactions to explain the variance of the response, indicating the OAK model is able to find low dimensional representation\footnotemark.}
	\label{fig:sobol}
\end{figure*}

\begin{figure*}[h!]
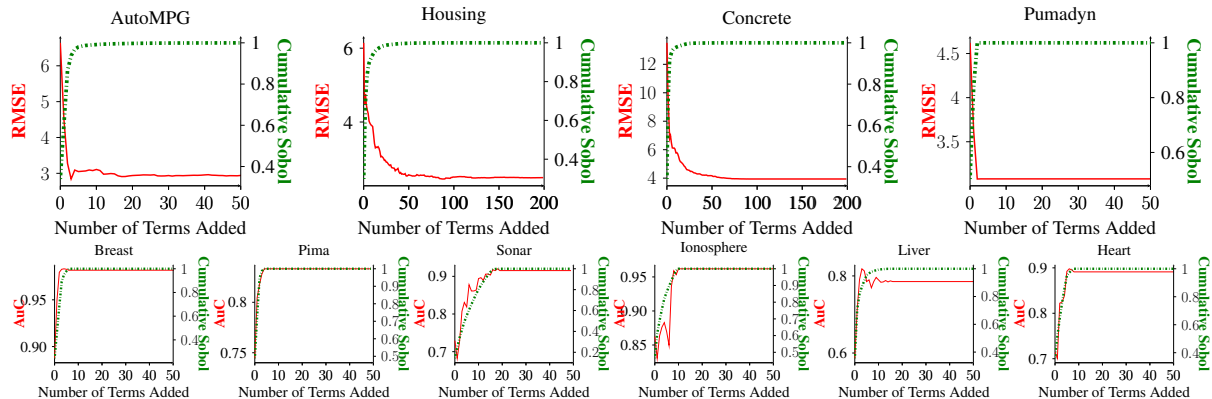

	\centering
	\vspace*{-0.1in}
	\begin{subfigure}[b]{0.23\textwidth}
		\input{tikz/UCI/autoMPG_cut_metric_and_sobol.tex}
	\end{subfigure} 
	\begin{subfigure}[b]{0.23\textwidth}
		\input{tikz/UCI/Housing_cut_metric_and_sobol.tex}
	\end{subfigure} 
	\begin{subfigure}[b]{0.23\textwidth}
		\input{tikz/UCI/concrete_cut_metric_and_sobol.tex}
	\end{subfigure} 
	\begin{subfigure}[b]{0.23\textwidth}
		\input{tikz/UCI/pumadyn_cut_metric_and_sobol.tex}
	\end{subfigure} 
	
	\vspace{-0.2in}
	\begin{subfigure}[b]{0.15\textwidth}
		\input{tikz/UCI/breast_cut_metric_and_sobol.tex}
	\end{subfigure} 
	\begin{subfigure}[b]{0.15\textwidth}
		\input{tikz/UCI/pima_cut_metric_and_sobol.tex}
	\end{subfigure} 
	\begin{subfigure}[b]{0.15\textwidth}
		\input{tikz/UCI/sonar_cut_metric_and_sobol.tex}
	\end{subfigure}
	\begin{subfigure}[b]{0.15\textwidth}
		\input{tikz/UCI/ionosphere_cut_metric_and_sobol.tex}
	\end{subfigure} 
	\begin{subfigure}[b]{0.15\textwidth}
		\input{tikz/UCI/liver_cut_metric_and_sobol.tex}
	\end{subfigure} 
	\begin{subfigure}[b]{0.15\textwidth}
		\input{tikz/UCI/heart_cut_metric_and_sobol.tex}
	\end{subfigure}
	\vspace{-0.2in}
	\caption{Model performance and cumulative Sobol index versus number of terms added ranked by the Sobol index. 
	We use test RMSE and area-under-the-curve (AuC) as the evaluation metric for regression problems (top) and classification problems (bottom) respectively. 
	Red solid lines represent test RMSE (top) and test AuC (bottom), green dashed lines represent cumulative (normalized) Sobol index.}
	\label{fig:sobol_plots}
\end{figure*}

\addtocounter{footnote}{-3} 
Our experiment procedure runs as follows: we plug the OAK kernel in the gpflow\footnote{\url{https://github.com/GPflow/GPflow}} package,
we then perform inference on regression problems with \verb+gpflow.GPR+ (or \verb+gpflow.SGPR+ for larger datasets); for classification tasks, we use \verb+gpflow.SVGP+ for inference We place a Gamma prior on the variance hyperparameters of the kernel, which are estimated using Maximum a Posterior (MAP). After learning the hyperparameters, we compute the Sobol index for each term including all orders of interactions up to the truncated order. Then we rank the importance of each term according to their Sobol
indices and investigate how many terms are needed to give competitive model performance. Details on the procedure can be found in Appendix
\ref{sec:oak_procedure}. 

We apply normalizing flows on all continuous features in our experiments before fitting the GP model, except for the Concrete dataset where the normalizing flow was not sufficient to transform the data and we have reverted to an empirical measure in this case. Empirically we have found that the model performance is similar with or without the normalizing flow, but the resulting model tends to be less parsimonious without the flow. More details and an ablation experimental study can be found in Appendix \ref{sec:nf}.

We validate our model on a range of experiments, including a set of regression and classification problems from datasets used in \citet{duvenaud2011additive} and additional UCI datasets, a large scale SUSY experiment and a Churn modelling problem. In our experiments we found OAK contains lower order terms without loss in predictive accuracy in contrast to  \citet{duvenaud2011additive} which finds higher order effects across a range of regression and classification problems. With OAK, only a small number of terms are needed in the model despite the large number of features available.

\stepcounter{footnote}\footnotetext{For all the classification datasets, the cumulative (normalized) Sobol indices for first order terms are found to be close to 1.}

\subsection{Baseline Experiments} \label{sec.duvenaud_experiments}
We first duplicate the experiments in \citet{duvenaud2011additive} where the number of instances and dimensionality of each dataset can be found in Appendix \ref{sec:uci_data}. We use five-fold cross-validation splits and compute test RMSE for regression and area-under-the-curve (AuC) errors for classification datasets. We use a GP with a squared exponential kernel as a baseline model to compare the  performance of OAK and the unconstrained additive GP model used in \citet{duvenaud2011additive}. For regression datasets, we set $\tilde{D} = D$; for classification problems, we set $\tilde{D} = 4$ except the Sonar dataset with $\tilde{D}=2$ for computational considerations. We found no significant differences in performance between different models, see more details in Appendix \ref{sec:UCI}.

Often one is interested in understanding how much each feature or interaction of features contribute in predicting the output. For example, one may ask how much does the 3rd order interaction term affect the response, or whether some feature is more important than others in explaining the response. We first plot the cumulative Sobol index for each order of interactions in Figure \ref{fig:sobol}, which is defined as the sum of Sobol indices for all terms in the same order. The results indicate that most datasets only need low order $(<3)$ interaction terms. 

Importantly, despite there being a large number of terms including all orders of interactions terms, we found that only a few terms are needed in the model to reach competitive performance. In Figure \ref{fig:sobol_plots} we plot model performance and cumulative Sobol as a function of the number of terms added, where the terms to add are ranked by their Sobol indices from highest to lowest. We report test RMSE and AuC for regression and classification problems respectively\footnote{\small We used empirical measure for Concrete dataset as its input feature distributions suggest.} . 

For each dataset with dimension $D$ and truncated maximum order of interaction $\tilde{D}$, the total number of terms is $\sum_{d=1}^{\tilde{D}} {D \choose d}$ (127 for autoMPG and 41448 for ionosphere datasets to give a sense of the scale, details in Appendix \ref{sec:number_of_terms}). We can observe the strong correlation between cumulative Sobol index and model performance. Only a few number of terms are needed before the model converges, indicating further terms add little value and OAK is able to find simple representations without loss of model performance. We further verify its parsimony by comparing the interaction order variance hyperparameter $\sigma_d^2$, see Appendix \ref{sec:relative_var}.

In particular, unlike in \citet{duvenaud2011additive} where an 8-dimensional interaction is required, we are able to reach the same model performance with only two first order terms and one second order term, which explain $>99\%$ variance in $f$. Due to the advantages of low-dimensional representation, we can visualise the decomposed functions with highest Sobol indices easily (Figure \ref{fig:pumadyn}). For completeness, we have also conducted experiments with the kernel $\prod_d (1 + \tilde{k}_d)$ used in \citet{duvenaud2011additive}, but using the constrained $\tilde{k}_d$. We found this kernel is harder to optimize and numerically less stable; the model performance is similar but the resulting model is less parsimonious (see Appendix \ref{sec:product_kernel}).

We conduct further experiments on an extensive range of benchmark datasets \cite{Salimbeni2018} with results displayed in Table \ref{table:additional_benchmarks}. 
We show summary statistics including the average, median and rank across the datasets.
For regression tasks we report test RMSE and log likelihood whereas for classification tasks we report test accuracy and log likelihood. 
We found the performance of OAK is on par or better compared with other methods. Detailed performance metrics on each dataset can be found in Appendix \ref{sec:additional_benchmarks}.
\begin{table*}
	\centering
	\small
	\begin{tabular}{l|l|lllllllllll}
    \toprule
     \multirow{4}{*}{Regression RMSE} &  Aggregation &        OAK &              Linear &         SVGP &          SVM &          KNN &          GBM &           AdaBoost &          MLP \\
    \midrule
    &      avg &        0.475 &            6.157 &        0.478 &        0.484 &        0.518 &        0.455 &        0.581 &        \textbf{0.445} \\
     &    median &        0.376 &            0.736 &        0.397 &        0.419 &        0.454 &        \textbf{0.343} &        0.580 &        0.361 \\
     &  avg rank &        3.583 &            6.625 &        4.083 &        4.208 &        4.958 &        \textbf{3.208} &        5.750 &        3.583 \\
    \midrule
    \multirow{3}{*}{Regression Log Likelihood}  & avg &          \textbf{-0.229} &        -0.946 &        -0.295 &        -0.585 &        -0.638 &        -0.652 &        -0.730 &         -0.891 \\
    & median &          \textbf{-0.409} &        -1.096 &        -0.512 &        -0.609 &        -0.738 &        -0.671 &        -0.875 &         -0.471 \\
    & avg rank &        \textbf{5.583} &         3.625 &         5.042 &         4.833 &         3.917 &         4.292 &         3.583 &          5.125 \\
   \midrule
   \multirow{3}{*}{Classification Accuracy} & avg &        \textbf{0.872} &        0.835 &        0.859 &        0.857 &        0.836 &        0.870 &        0.859 &        0.863 \\
    & median &          0.898 &        0.832 &        0.864 &        0.850 &        0.863 &        \textbf{0.900} &        0.892 &        0.873 \\
    & avg rank &       \textbf{5.569} &        4.224 &        4.741 &        4.500 &        2.983 &        5.224 &        4.207 &        4.552 \\
    \midrule
    \multirow{3}{*}{Classification Log Likelihood} & avg &          \textbf{-0.267} &        -0.338 &        -0.291 &        -0.306 &        -0.899 &        -0.283 &        -0.459 &        -0.306 \\
    & median &       -0.280 &        -0.389 &        -0.307 &        -0.352 &        -1.088 &        \textbf{-0.256} &        -0.584 &        -0.362 \\
    & avg rank &       5.862 &         4.276 &         \textbf{5.931} &         4.690 &         2.138 &         5.379 &         2.897 &         4.828 \\
\bottomrule
    \end{tabular}
    
\caption{Experimental results on additional benchmark datasets. Average results over 24 regression datasets shown in
 terms of test RMSE and log likelihood (top two blocks). Average results  over 29 classification datasets 
 shown in terms of accuracy and log likelihood (bottom two blocks). Higher is better except for RMSE. 
 SVGP=Stochastic Variational GP, using GPflow \citep{hensman2015scalable}; SVM=Support Vector Machine, 
 KNN=K-nearest-neighbours,  GBM=Gradient Boosting Machine, MLP=Multi-layer Perceptron (all using Scikit-learn defaults). Results compiled using the Bayesian 
 Benchmarks repo \citep{Salimbeni2018}. Full results are shown in Appendix \ref{sec:additional_benchmarks}.}
\label{table:additional_benchmarks}
\end{table*}

\subsection{SUSY Classification} \label{sec.susy}
In the next experiment we tackle a large-scale binary classification problem. The super-symmetric (SUSY\footnote{\small\href{https://archive.ics.uci.edu/ml/datasets/SUSY}{archive.ics.uci.edu/ml/datasets/SUSY}}) 
dataset contains 5 million instances with 8 low level kinematic properties, where the task is to predict whether a signal process produces super-symmetric particles or not. We truncate $\tilde{D} = 2$ for computational consideration. 

\begin{figure}[h!]
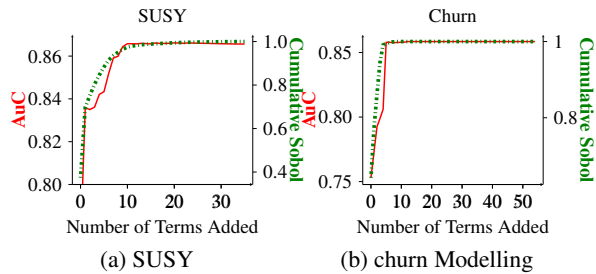

	\centering
	\vspace{-0.1in}
	\begin{subfigure}[b]{0.22\textwidth}
		\input{tikz/susy/auc_and_sobol.tex}
		\vspace*{-0.2in}
		\caption{SUSY}
		\label{fig:susy_sobol}
	\end{subfigure} 
	\begin{subfigure}[b]{0.22\textwidth}
		\input{tikz/churn/auc_and_sobol.tex}
		\vspace*{-0.2in}
		\caption{churn Modelling}
		\label{fig:churn_sobol}
	\end{subfigure} 
	\vspace{-0.1in}
	\caption{AuC as a function of number of terms added ranked by their Sobol indices for the SUSY (left) and Churn modelling (right) experiments. Red solid lines and green dashed lines represent test AuC and cumulative (normalized) Sobol respectively.}
\end{figure}

We use the same training-test split as in \citet{dutordoir2020sparse}. We fit a sparse variational GP (SVGP) model with OAK and optimize the variational parameters and hyperparameters with natural gradients and Adam respectively. Number of inducing points and mini batch size are set to be 800 and 1024 respectively.

\begin{figure*}[!h]
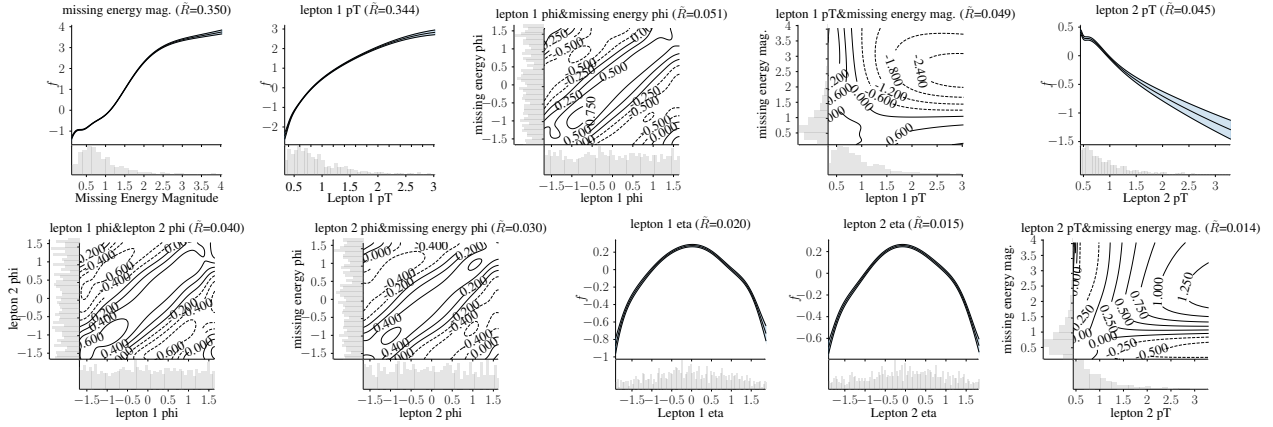

	\centering
	\vspace*{-2mm}
	\begin{subfigure}[t]{0.16\textwidth}
		\input{tikz/susy/missing_energy_mag.tex}
	\end{subfigure} 
	\begin{subfigure}[t]{0.16\textwidth}
		\input{tikz/susy/lepton_1_pT.tex}
	\end{subfigure} 
	\begin{subfigure}[t]{0.215\textwidth}
		\input{tikz/susy/lepton_1_phi_missing_energy.tex}
	\end{subfigure} 
	\begin{subfigure}[t]{0.215\textwidth}
		\input{tikz/susy/lepton_1_pT_missing_energy.tex}
	\end{subfigure} 
	\begin{subfigure}[t]{0.16\textwidth}
		\input{tikz/susy/lepton_2_pT.tex}
	\end{subfigure} 
	
	\begin{subfigure}[b]{0.215\textwidth}
		\input{tikz/susy/lepton_1_phi_lepton_2_phi.tex}
	\end{subfigure} 
	\begin{subfigure}[b]{0.215\textwidth}
		\input{tikz/susy/lepton_2_phi_missing_energy_phi.tex}
	\end{subfigure} 
	\begin{subfigure}[b]{0.16\textwidth}
		\input{tikz/susy/lepton_1_eta.tex}
	\end{subfigure} 
	\begin{subfigure}[b]{0.16\textwidth}
		\input{tikz/susy/lepton_2_eta.tex}
	\end{subfigure} 
	\begin{subfigure}[b]{0.215\textwidth}
		\input{tikz/susy/lepton_2_pT_missing_energy_mag.tex}
	\end{subfigure} 
	\vspace*{-1mm}
	\caption{Decomposition of top 10 important functions for SUSY dataset, ranked by their Sobol indices. Blue shaded area represents uncertainties with two standard deviation. Grey shaded area represent histograms of input features. $\tilde{R}$ in the brackets denote (normalized) Sobol index. Missing energy magnitude and lepton 1 pT are the two most important features which explain $\approx 70\%$ of the variance in the model $f$, and they both have a positive impact where a signal process is more likely to produces semi-symmetric particles when missing energy magnitude and lepton 1 pT are high. }
	\label{fig:susy_plots}
\end{figure*}

Model performance are reported in Table \ref{table:susy_churn} where the OAK model achieves similar or better performance compared with other deep learning models.  Top 10 important functional components are displayed in Figure \ref{fig:susy_plots}, which contain five first order terms and five second order terms. In particular, a signal process is more likely to produce super-symmetric particles if there is higher missing energy magnitude; higher lepton 1 pT or lower lepton 2 pT. For lepton 1 eta or lepton 2 eta, the probability first increases and then decreases with increasing values of eta. In Figure \ref{fig:susy_sobol} we can observe that with these 10 terms, we are able to reach  the optimal AuC and capture $96\%$ of the variance in $f$. This further shows that the OAK model is able to reach competitive performance while having a simple, interpretable representation.

\begin{figure*}[h!]
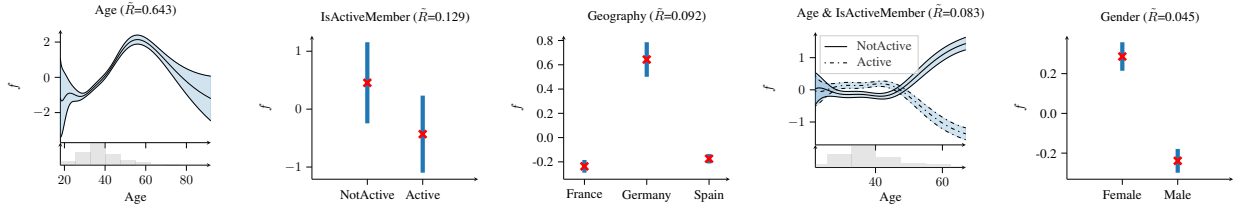

	\vspace*{-2mm}
	\centering
	\begin{subfigure}[b]{0.19\textwidth}
		\input{tikz/churn/age.tex}
	\end{subfigure} 
	\begin{subfigure}[b]{0.19\textwidth}
		\input{tikz/churn/active.tex}
	\end{subfigure} 
	\begin{subfigure}[b]{0.19\textwidth}
		\input{tikz/churn/geography.tex}
	\end{subfigure} 
	\begin{subfigure}[b]{0.19\textwidth}
		\input{tikz/churn/age_active.tex}
	\end{subfigure} 
	\begin{subfigure}[b]{0.19\textwidth}
		\input{tikz/churn/gender.tex}
	\end{subfigure} 
  \vspace*{-0.1in}
	\caption{Decomposition of top 5 important functions for Churn dataset, ranked by their Sobol indices. Blue shaded area represents uncertainties with two standard deviation. $\tilde{R}$ in the brackets denote (normalized) Sobol index. Age is the most important feature in predicting whether a customer leaves the bank or not, typically as one gets older (but younger than 55), he/she is more likely to leave the bank. Non-active members, female customers and German customers are more likely to leave the bank compared to their counterparts. The interaction between age and whether a customer is active also contributes to the probability: older non-active customers and younger active customers are more likely to churn.}
	\label{fig:churn_plots}
		\vspace*{-0.1in} 
\end{figure*}

\begin{table}
	\tiny
	\begin{center}
		\begin{tabular}{ |c|c|c|c| } 
			\hline
			\multicolumn{2}{|c|}{SUSY}  & \multicolumn{2}{|c|}{Churn}   \\ \hline 
			Method & AuC & Method & AuC \\ \hline 
			BDT* & 0.850 $\pm$  0.003  &XGBoost&0.853 $\pm$  0.008 \\ 
			NN* & 0.867 $\pm$  0.002  &MLP*&0.846 $\pm$  0.013\\ 
			$\text{NN}_{\text{dropout}}$* & 0.856 $\pm$  0.001  &Sparse MLP*&0.828 $\pm$  0.007\\ 
			SVGP(SE)* & 0.852 $\pm$  0.002  &TabTransformer*&0.856 $\pm$  0.005\\ 
			VISH* & 0.859 $\pm$  0.001  &TabNet*&0.785 $\pm$  0.024\\ 
			OAK & 0.865 $\pm$  0.0004 &OAK&0.856 $\pm$  0.009\\
			\hline
		\end{tabular}
	\end{center}
	\vspace*{-0.1in} 
	\caption{Performance comparison for SUSY (left) and Churn modelling (right). The mean AuC is reported with one standard deviation, with 5 repetitions (SUSY) and 5 cross-validation splits (Churn) respectively. Larger is better. Results with * are quoted from \citet{dutordoir2020sparse} and \citet{huang2020tabtransformer}.}
	\label{table:susy_churn}
\end{table}
\subsection{Churn Modelling} \label{sec.churn}
Next we look at Churn Modelling problem available from Kaggle\footnote{\small\href{https://www.kaggle.com/shrutimechlearn/churn-modelling}{www.kaggle.com/shrutimechlearn/churn-modelling}}.
This data set contains details of a bank's customers where the goal is to predict whether the customer leaves the bank or continues to be a customer. There are 10 features including a mix of continuous and categorical variables such as age, gender, credit score, etc.. We truncate the maximum order of interactions to be 2 for computational consideration. 
We compare model performance with XGBoost, MLP, TabNet and TabTransformer with the same training-test split (4:1) as in \citet{huang2020tabtransformer}. Test AuC are reported in Table \ref{table:susy_churn}. We outperform or are as accurate as all of the baseline models with increased interpretability. 

In Figure \ref{fig:churn_sobol} we observe that only $\approx 5$ terms are needed for the model to achieve optimal performance. We plot the top 5 important features/interactions in Figure \ref{fig:churn_plots} based on Sobol indices, which contain four first order terms and one interaction term between Age and IsActiveMember with the following insights: Age is the most important feature in predicting whether a customer leaves the bank, and generally the older a person is, more likely they will leave the bank; more active members are less likely to leave; German people are more likely to leave compared with French and Spanish; women are more likely to leave compared with men. The interaction between Age and IsActiveMember says that for less active customers, older people are more likely to leave the bank whereas for active members, older customers are more likely to stay.

\section{Conclusion}
\label{sec:conclusion}
In this work, we have proposed a Gaussian process model with orthogonal additive kernel (OAK) that enables inference of low-dimensional representations that are identifiable and interpretable. The resulting model has an analytic form for the Sobol indices which can be used to rank importance of features and interactions. We have shown that the OAK model allows inference of low-dimensional representations whilst achieving state-of-the-art predictive performance on a range of both regression and classification tasks. We are surprised to find out all the datasets we have experimented with can be modelled using low dimensional functions.

One limitation of our work is that we implicitly assumed independence between input features and independent, identically distributed Gaussian noise. Future work can extend our approach to non-independent input features and examine the effect of heteroscedastic noise using latent variable models. Another interesting direction of work is to extend OAK to Bayesian optimization and experimental design leveraging the inferred low-order representation. 

\section*{Acknowledgement}
The authors would like to thank Nicolas Durrande for insightful discussions, especially around the orthogonality construction. 
Thanks to George Michailidis, Dominic Richards and Fran\c{c}ois-Xavier Aubet
for valuable feedback on the manuscript, helpful discussions on the connections to Sobol indices, and advice on ICML rebuttals. We thank Vincent Dutordoir and Stefanos Eleftheriadis for sharing code and for help with natural gradient and Adam optimization for SVGP models.  Finally, thanks to David Duvenaud, Hannes Nickisch and Carl Rasmussen whose open code and thoughtful paper inspired this work.

\bibliography{ref}
\bibliographystyle{icml2022}

\newpage
\appendix
\onecolumn
\onecolumn
\section{Newton-Girard Method for Computing the Interacting Kernel}
\label{sec:newton_girard_alg}
\begin{algorithm}
	\begin{algorithmic}
	\STATE  {\bfseries Input:} input dimension $D$
	\STATE{\bfseries Input:} maximum interaction order $\tilde{D}$
	\STATE{\bfseries Input:} base kernels $k_d(\cdot, \cdot), d=1\ldots D$
	\STATE{\bfseries Input:} order variances $\sigma_l, l=0\ldots \tilde{D}$
	\STATE{\bfseries Data:} input data $\mathbf X$
	\STATE{\bfseries Output:} kernel matrix $\mathbf K$
	\FOR{$d=1\ldots D$}
	\STATE{
		$\mathbf K_d[i, j] = k_d(x_{i,d}, x_{j, d})$
	}
    \ENDFOR
	\FOR{$\ell=0\ldots \tilde{D}$}
	\STATE{
		$\mathbf S_\ell = \sum_{d=1}^D \mathbf K_d ^ \ell$
	}
	\ENDFOR
	\STATE$\mathbf E_0 = \mathbf 1^{[N, N]}$
	
	\FOR{$\ell$=1\ldots$\tilde{D}$}
	\STATE{
		$\mathbf E_\ell = \frac{1}{\ell}\sum_{k=1}^{\ell}(-1) ^{k - 1} \mathbf E_{\ell - k} \odot \mathbf S_k$
	}
	\ENDFOR
	\STATE$\mathbf K = \sum_{\ell=0}^{\tilde{D}} \sigma_\ell \times \mathbf E_\ell$
	\end{algorithmic}
	\caption{Newton-Girard method for computing the interacting kernel}
	\label{alg:newton_girard}
\end{algorithm}

\section{Orthogonality in Higher Dimension}
\label{sec:orthogonality_product}
For higher order terms, recall OAK uses the product of constrained kernels (equation (\ref{eq:product_kernel})):
\begin{align}
\tilde{k}_u(x,x') = \prod_{i \in u} \tilde{k}_i(x_i,x_i').
\end{align}
We show the product of constrained kernel satisfies the orthogonality constraint in higher dimensions, i.e.,$\forall i \in u$,
\begin{align} 
\int_{\mathcal{X}_i} f_u(\x_u)p_i(x_i)dx_i = 0
\end{align}
where each functional component $f_u$ has kernel $k_u$.
\begin{proof}
	By construction, for each function $i$ with constrained kernel $\tilde{k}_i$, $f_i$ satisfies the orthogonality constraint $S_i := \int_{\mathcal{X}_i} f_i(x_i)p_i(x_i)dx_i = 0$ (equation (\ref{eq:orthogonal_constraint})), which implies that:
	\begin{align}
	\E_{f_i}[S_i] = 0, \;\;\;\; \mathbb{V}_{f_i}[S_i] = \int_{\mathcal{X}_i} \tilde{k}_i(\x_i, \x_i)p_i(x_i)dx_i = 0.
	\end{align}
	To prove $\int_{\mathcal{X}_i} f_u(\x_u)p_i(x_i)dx_i = 0$, it is sufficient to prove the mean and variance of $\int_{\mathcal{X}_i} f_u(\x_u)p_i(x_i)dx_i $ with respect to $f_u$ is zero. Since we assume $f_u$ has zero mean, the mean $\E_{f_u}\left[\int_{\mathcal{X}_i} f_u(\x_u)p_i(x_i)dx_i\right] = 0.$ The variance
		\newpage
	\begin{align}
	\mathbb{V}_{f_u}\left[\int_{\mathcal{X}_i} f_u(\x_u)p_i(x_i)dx_i\right] &= 
	\int_{\mathcal{X}_i} \E_{f_u}[f_u(\x_u)^2]p_i(x_i)dx_i  \nonumber \\
	&=\int_{\mathcal{X}_i} k_u(\x_u,\x_u)p_i(x_i)dx_i \label{eq:var_constraint} \\
	&=\prod_{j\neq i} k_j(x_j,x_j) \int_{\mathcal{X}_i} k_i(\x_i,\x_i)p_i(x_i)dx_i = 0. \nonumber
	\end{align}
\end{proof}

\section{Constrained Categorical Kernel}
\label{sec:coregional}
For categorical input features, we can model $f$ with the categorical kernel as in  \citet{coregionalkernel}, which is constructed by a positive definite matrix $A$ such that the categorical kernel $k(i,j) = A_{ij}$ where
\begin{align}
A = WW^\top + \text{Diag}(\kappa).
\end{align}
The orthogonality constraint we put on $f$ is $\int f(x)p(x) dx = 0$. Let $\w$ be the vector of probability measure of the input feature, i.e., $\mathbb{P}(x = i) = w_i$ for $i = 1, \cdots, M$.
Define 
\begin{align}
B := A - \frac{A\w(A\w)^\top}{\w^\top A\w},
\end{align}
we claim the kernel with $\tilde{k}(i,j) = B_{ij}$  is the constrained categorical kernel. To see this, it is enough to show $\mathbb{E}_{p(i,j)}[k(i,j)] = 0$ as shown in (\ref{eq:var_constraint}):
	\begin{align}
	\mathbb{E}_{p(i,j)}[\tilde{k}(i,j)] = \sum_{i=0}^M \sum_{j=0}^M \tilde{k}(i,j)w_iw_j = \w^\top A\w - \w^\top \left(\frac{A\w\w^\top A}{\w^\top A\w}\right)\w = 0.
	\end{align}

\section{Constrained Squared Exponential Kernel}
\label{sec:input_measure}
\subsection{Gaussian Measure}
We prove the constrained squared exponential kernel takes the form in (\ref{eq:crbf}) when the input feature has Gaussian density.
Assume squared exponential kernel with lengthscale $l$ and variance $\sigma^2$, and Gaussian measure $p(x) \sim \mathcal{N}(\mu,\delta^2)$. Denote $S:= \int f(x)p(x)dx$, by (\ref{eq:durrande_kernel}) we need to calculate:
\begin{align}
\E_f[Sf(a)] &= \int \sigma^2 p(x) \exp\left(-\frac{(x-a)^2}{2l^2}\right) dx    \\
&= \int \frac{\sigma^2}{\sqrt{2\pi \delta^2}} \exp{\left(-\frac{(x-a)^2}{2l^2}\right)} \exp{\left(-\frac{(x-\mu)^2}{2\delta^2}\right)} dx \\
&= \int \sigma^2 \sqrt{2\pi l^2} \mathcal{N}(x; a,l^2) \mathcal{N} (x; \mu,\delta^2) dx \\
&= \sigma^2\sqrt{\frac{l^2}{l^2 + \delta^2}} \exp{\left(-\frac{(a-\mu)^2}{2(l^2+\delta^2)}\right)},
\end{align}
and
\begin{align}
\E_f[S^2] &= \int\int p(x) p(x') k(x,x') dx dx' \\
&= \int \sigma^2\sqrt{\frac{l^2}{l^2 + \delta^2}} \exp{\left(-\frac{(x-\mu)^2}{2(l^2+\delta^2)}\right)} p(x) dx \\
&= \sigma^2 \sqrt{\frac{l^2}{l^2 + 2\delta^2}}
\end{align}
where the last equality follows from completing the square. 

\subsection{Mixture of Gaussian Measure}
We can extend the Gaussian density assumption to other input distributions such as mixture of Gaussians. Suppose a fixed number of clusters $K$:
\begin{equation}
p(x) = \sum_{k=1}^K w_k N(\mu_k, \delta_k)
\end{equation}
where $\mu_k, \delta_k$ are the mean and variance of each cluster.

From (\ref{eq:durrande_kernel}), two expectations need to be calculated to compute the constrained kernel, the variance $\E_f[S^2]$ and the covariance $\E_f[S f(x)]$ can be computed as
\begin{align}
&\E_f[S^2] =  \sum_{i=1}^K \sum_{j=1}^K w_i w_j  l N(\mu_i | \mu_j, l^2 + \delta_i + \delta_j) (2 \pi)^{1/2},\\
&\E_f[S f(x)] = \sum_{k=1}^K lw_k N(x | \mu_k, \delta_k + l^2) (2\pi)^{1/2}  
\end{align}
where $l$ is the kernel lengthscale parameter and we have assumed unit kernel variance parameter for simplicity.

\section{Constrained Kernel under Empirical Measure}
\label{sec:empirical_measure}
When input densities are far from (mixture of) Gaussian distributions, or categorical kernel is not appropriate, or one wants to use other kernels, we can use the empirical measure $p(x) = \sum_{i=1}^M w_i\mathbbm{1}_{x=x_i}$, where $\{x_i\}_{i=1}^M$ are the locations of the feature and $\{w_i\}_{i=1}^M$ are the associated weights. We can approximate (\ref{eq:durrande_kernel}) with
\begin{align}
\E_{f}[S f(\cdot)] \approx \sum_{i=1}^M w_i k(x,x_i),  \;\;\;\;
\E_{f}[S^2] \approx  \sum_{i=1}^M \sum_{j=1}^M w_iw_j k(x_i,x_j).
\end{align} 

\section{Normalizing Flow}
\label{sec:normalising_flow}
Specifically, let $\{x^i\}_{i=1}^N$ be the data for feature $x$ with unknown underlying density $p_x(x)$. We apply a sequence of $K$ bijective functions to obtain the transformed features $z$:
\begin{align}
z = f_K \circ f_{K-1} \circ f_1(x) := g(x).
\end{align}
The density of $z$ can be calculated as:
\begin{align}
p_z(z) = \frac{1}{g'(x)}p_x(x) \approx \frac{1}{N}\sum_{i=1}^N  \frac{1}{g'(x^i)} \mathbbm{1}_{x=x^i},
\end{align}
where $g'$ denotes the derivative. We would like $z$ to be as close to standard Gaussian distributed as possible, denote $p(z)$ to be $\mathcal{N}(0,1)$, we minimize the KL-divergence: 
\begin{align}
\textit{KL}(p_z(z) || p(z)) &= \E_{p_z(z)}\left[\log \frac{p_z(z)}{p(z)}\right] \\
&\approx \frac{1}{N} \sum_{i=1}^N \left[\log \frac{p_z(z^i)}{p(z^i)}\right] \\
& =\frac{1}{N} \sum_{i=1}^N \left((z^i)^2 - \log g'(x^i)\right) + C
\end{align}
where $C$ is some constant and we approximated $p_z$ with its empirical distribution. The parameters of $g$ are then learnt by minimizing this KL divergence.

\section{Sobol Indices}
\label{sec:sobol_appendix}
Recall the normalized Sobol index for the posterior mean of $f_u$ for $u \in [D] $ is: 
\begin{equation}
\tilde{R}_u = \frac{\mathbb{V}_\x [ m_u(\x) ]} { \mathbb{V}_\x [ m(\x)] }.
\end{equation}
The posterior mean GP with component $u$ is:
\begin{equation}
m_u(x) = \sigma_{|u|}^2\left( \odot_{i \in u} k_i(x_i, \X_i) \right) K(\X,\X)^{-1} y \label{eq.postmean}
\end{equation}
where $K(\X,\X)$ denotes the training input covariance across all inputs, $y$ denotes the $n \times 1$ vector of output observations, $\sigma_{|u|}^2$ is the associated variance parameter for $|u|$-th order interaction and $ \odot$ denotes element-wise multiplication. A similar formula for sparse GP can also be obtained. The posterior variance with respect to the input is therefore
\begin{align}
\mathbb{V}_x [ m_u(x) ] &= \mathbb{V}_x \left[  \sigma_{|u|}^2  \left( \odot_{i \in u} k_i(x_i, \X_i)  \right) K(\X,\X)^{-1}  y  \right] \nonumber \\
&= \sigma_{|u|}^4 y ^\top K(\X,\X)^{-1} \cov  \left[  \odot_{i \in u} k_i(x_i, \X_i) \right] K(\X,\X)^{-1}  y \nonumber\\
&= \sigma_{|u|}^4 y^\top K(\X,\X)^{-1}   \odot_{i \in u} \left( \int k_i(x_i, \X_i) k_i(x_i, \X_i)^\top dp_i(x_i) \right) K(\X,\X)^{-1}  y.
\label{eq:sobol_appendix}
\end{align}
In case of 1) constrained squared exponential kernel and a Gaussian measure or 2) binary/categorical kernel with discrete measure, the integral is tractable and can be computed analytically. 

\subsection{Sobol Index for Constrained Squared Exponential Kernel}
To compute the Sobol index, we need to compute the integral in (\ref{eq:sobol_appendix}). Dropping subscript $i$ for simplicity, assume one-dimensional feature $X$, squared exponential base kernel with lengthscale $l$ and variance $\sigma^2$: $k(x,x') = \sigma^2\exp\left(-\frac{1}{2l^2}(x-x')^2\right)$ and Gaussian input density $p(x) = \mathcal{N}(\mu,\delta^2)$, recall the constrained squared exponential kernel $\tilde{k}$ is :
\begin{align}
\tilde{k}(x,x') &:= k(x,x') - \frac{\sigma^2 l \sqrt{l^2 + 2\delta^2 }}{l^2 + \delta^2}\exp\left(-\frac{1}{2(l^2 + \delta^2)}((x-\mu)^2 + (x'-\mu)^2)\right) \\
&:= k(x,x') - \hat{k}(x,x')
\end{align}
where
\begin{align}
\hat{k}(x,x') := \frac{\sigma^2 l \sqrt{l^2 + 2\delta^2 }}{l^2 + \delta^2}\exp\left(-\frac{1}{2(l^2 + \delta^2)}((x-\mu)^2 + (x'-\mu)^2)\right) .
\end{align}
Denote $a=X_p$, $b=X_q$ respectively, The $(p,q)$-entry is of $\int \tilde{k}(x, X) \tilde{k}(x, X)^\top dp(x)$ in (\ref{eq:sobol_appendix}) is therefore
\begin{align}
\int p(x) \tk(x,a) \tk(x,b) dx &= \int p(x)  k(x,a)  k(x,b) dx \label{eq:eq1} \\
&-  \int p(x)  k(x,a) \hat{k}(x,b) dx \label{eq:eq2} \\
&- \int p(x)  \hat{k}(x,a) k(x,b) dx \label{eq:eq3} \\
&+ \int p(x)  \hat{k}(x,a)  \hat{k}(x,b) dx \label{eq:eq4}.
\end{align} 
We compute each of the term in following subsections.

\subsubsection{equation (\ref{eq:eq1})}
\begin{align*}
\int p(x)  k(x,a)  k(x,b) dx &= \int p(x) \sigma^4 \exp\left(-\frac{1}{2l^2}\left((x-a)^2 + (x-b)^2\right)\right)dx \\
&= \sigma^4 \int p(x) \exp\left(-\frac{1}{2l^2}(2x^2 -2(a+b)z + a^2 + b^2)\right)dx \\
&= \sigma^4 \exp\left(-\frac{1}{2l^2}(x^2 + y^2)\right) 
\int p(x) \exp\left(-\frac{1}{l^2}(x^2 - (a+b)z)\right)dx \\
&= \sigma^4 \exp\left(-\frac{1}{2l^2}(a^2 + b^2)\right)  \exp\left(-\frac{1}{l^2}\left(\frac{a+b}{2}\right)^2\right)
\int p(x) \exp\left(-\frac{1}{l^2}\left(x-\frac{a+b}{2}\right)^2\right)dx.
\end{align*}
Note
\begin{align*}
\int p(x) \exp\left(-\frac{1}{l^2}\left(x-\frac{a+b}{2}\right)^2\right)dx &= \sqrt{\pi l^2} \int \mathcal{N}(z;\mu,\delta^2)\mathcal{N}\left(x;\frac{a+b}{2},\frac{l^2}{2}\right)dx \\
&= \frac{l}{\sqrt{2\delta^2 + l^2}}\exp\left(-\frac{1}{2\delta^2 + l^2}\left(\mu - \frac{a+b}{2}\right)^2\right).
\end{align*}
Hence
\begin{align*}
\int p(x)  k(x,a)  k(x,b) dx = \frac{\sigma^4 l}{\sqrt{2\delta^2 + l^2}}\exp\left(-\frac{1}{4l^2}(a-b)^2\right)
\exp\left(-\frac{1}{2\delta^2 + l^2} \left(\mu-\frac{a+b}{2}\right)^2\right).
\end{align*}
\subsubsection{equation (\ref{eq:eq2})}
\begin{align*}
\int p(x)  k(x,a) \hat{k}(x,b) dx = \frac{\sigma^4 l \sqrt{l^2 + 2\delta^2}}{l^2 + \delta^2} \exp\left(-\frac{1}{2(l^2 + \delta^2)}(b-\mu)^2\right)
\int p(x) \exp\left(-\frac{(x-a)^2}{2l^2} - \frac{(x-\mu)^2}{2(l^2 + \delta^2)} \right)dx.
\label{eq:tmp}
\end{align*}
Note
\begin{align*}
\int p(x) \exp\left(-\frac{(x-a)^2}{2l^2} - \frac{(x-\mu)^2}{2(l^2 + \delta^2)} \right)dx  &= \int p(x) \exp\left(-\frac{1}{2M^{-1}}(x-c)^2 + C \right) dx \\
&=\sqrt{2\pi M^{-1}}\exp\left(-\frac{C}{2}\right)\int \mathcal{N}(x; \mu,\delta^2) \mathcal{N}(x;c,M^{-1}) dx \\
&= \frac{1}{\sqrt{\delta^2M + 1}}\exp\left(-\frac{C}{2}\right)\exp\left(-\frac{1}{2(\delta^2 + M^{-1})}(c-\mu)^2\right),
\end{align*}
where
\begin{align*}
M := \frac{1}{l^2} + \frac{1}{l^2 + \delta^2}, \;\;\;\;\;
c := M^{-1}\left(\frac{\mu}{l^2 + \delta^2} + \frac{a}{l^2}\right) \;\;\;\;\;
C := \frac{a^2}{l^2} + \frac{\mu^2}{l^2 + \delta^2} - c^2M.
\end{align*}
Hence, 
\begin{align*}
\int p(x)  k(x,a) \hat{k}(x,b) dx = \frac{\sigma^4l\sqrt{l^2 + 2\delta^2}\exp(-C/2)}{(l^2 + \delta^2)\sqrt{\delta^2M + 1}}
\exp\left(-\frac{1}{2(l^2 + \delta^2)}(b-\mu)^2\right)
\exp\left(-\frac{1}{2(\delta^2 + M^{-1})}(c-\mu)^2\right).
\end{align*}
\subsubsection{equation (\ref{eq:eq3})}
By symmetry, this is straight-forward by interchanging $a$ and $b$ in (\ref{eq:eq2}).

\subsubsection{equation (\ref{eq:eq4})}
\begin{align*}
\int p(x)  \hat{k}(x,a)  \hat{k}(x,b) dx = \frac{\sigma^2 l^2 (l^2 + 2\delta^2)}{(l^2 + \delta^2)^2}
\exp\left(-\frac{(a-\mu)^2 + (b-\mu)^2}{2(l^2 + \delta^2)}\right) 
\int p(x) \exp\left(-\frac{(x-\mu)^2}{(l^2 + \delta^2)} \right)dx.
\end{align*}
Note
\begin{align*}
\int p(x) \exp\left(-\frac{1}{(l^2 + \delta^2)}(x-\mu)^2 \right)dx &= 
\sqrt{\pi(l^2+\delta^2)} \int \mathcal{N}(x;\mu,\delta^2) \mathcal{N}\left(x;\mu,\frac{l^2 + \delta^2}{2}\right) dx 
= \sqrt{\frac{l^2 + \delta^2}{l^2 + 3\delta^2}}.
\end{align*}
Hence
\begin{align*}
\int p(x)  \hat{k}(x,a)  \hat{k}(x,b) dx = \frac{\sigma^4l^2(l^2 + 2\delta^2)\sqrt{l^2 + \delta^2}}{(l^2 + \delta^2)^2\sqrt{l^2 + 3\delta^2}} \exp\left(-\frac{1}{2(l^2 + \delta^2)}((a-\mu)^2 + (b-\mu)^2)\right).
\end{align*}

\subsection{Sobol for Empirical Measure}
\label{sec:sobol_empirical}
Assume one-dimensional feature $x$ with empirical measure $p(x) = \sum_{i=1}^M w_i \textbf{1}_{x=x_{i}}$ where $M$ is the number of distinct empirical locations, $w_i$ are the (normalized) empirical weights, $x_i$ are the empirical locations for $i = 1, \cdots, M$. We can approximate the integral in (\ref{eq:sobol_appendix}) with
\begin{align}
\int k(\textbf{X},x) k(\textbf{X},x)^\top dp(x) 
\approx  \sum_{i=1}^M w_i 
k(\textbf{X},x_i) k(\textbf{X},x_i)^\top. 
\end{align}

\subsection{Proof of FANOVA for OAK}
\label{sec:anova_proof}
We show in this section that under the asumption that input features are independent, OAK results in the FANOVA decomposition, i.e., for each $u \subseteq [D]$, $f_u$ with $\tilde{k}_u$ satisfies that
\begin{align}
f_u(\x) = \int_{\mathcal{X}_{-u}} \left(f(\x) - \sum_{v \subset u} f_v(\x_v)\right) dP(\x_{-u}).
\end{align}
\begin{proof}
	The right hand side writes
	\begin{align}
	\int_{\mathcal{X}_{-u}} \left(f(\x) - \sum_{v \subset u} f_v(\x_v)\right) dP(\x_{-u}) &= 
	\int_{\mathcal{X}_{-u}} \left(f_u(\x) + \sum_{v \nsubseteq u} f_v(\x_v)\right) dP(\x_{-u})  \nonumber \\
	&= f_u(\x_u) + \sum_{v \nsubseteq u} \int_{\mathcal{X}_{-u}} f_v(\x_v) dP(\x_{-u}).
	\label{eq:anova_integral}
	\end{align}
For each $v \nsubseteq u$, if $j \in [D]\setminus u$, then $j \in v$. It follows from Appendix \ref{sec:orthogonality_product} that 
\begin{align*}
	\int_{\mathcal{X}_j} f_v(\x_v) dP(\x_j) = \int_{\mathcal{X}_j} f_v(\x_v) p_j(x_j) dx_j =0.
\end{align*}
Under the assumption that input features are independent, $P(\x_{-u})$ factorizes and the integral in equation (\ref{eq:anova_integral}) is 0.
\end{proof}

\subsection{Invariance of Sobol under Bijective Transformation}
\label{sec:invariance_sobol}
Let $Z \sim \mathcal{N}(0,1)$, and suppose $X$ is a transformation of $Z$ such that  $z = g(x)$ where $g$ is an invertible function. First note the density of $X$ is  
\begin{align}
p(x) = \mathcal{N}(g(x)|0,1)\left|\frac{dg(x)}{dx}\right |.
\end{align}
One can rewrite a function of $x$ as a function of $z$, suppose $f(x) = h(z) = h(g(x))$, the Sobol index for $x$ can be calculated as
\begin{align}
R &= \int f^2(x)p(x)dx \\
&= \int_{-\infty}^\infty h^2(g(x))p(x)dx\\
&= \int_{-\infty}^\infty h^2(g(x))p(x)\left|\frac{dg(x)}{dx}\right |^{-1}dz \\
&=  \int_{-\infty}^\infty h^2(g(x))\mathcal{N}(g(x)|0,1) dz\\
&=  \int_{-\infty}^\infty h^2(z)\mathcal{N}(z|0,1) dz,
\end{align}
which is the Sobol index for $Z$.

\section{OAK Method Summary}
\label{sec:oak_procedure}
Choose a truncation order for the model, $\tilde{D}$.
\vspace{-0.2cm}
\begin{enumerate}
	\itemsep0em 
	\item For each input dimension, a kernel is assigned:
	\begin{enumerate}
		\item Continuous features are assigned constrained squared exponential kernels, and transformed through a normalizing flow to ensure Gaussian input density.
		\item Discrete features are assigned a constrained binary or categorical kernel (see Appendix \ref{sec:coregional}).
	\end{enumerate}
	\item Fit a Gaussian process model with OAK defined in Section \ref{sec:orthogonality} and the Newton Girard Trick in Algorithm \ref{alg:newton_girard}.
	\begin{enumerate}
		\item For small ($N< 1000$) regression datasets, we use Exact Gaussian Process regression.
		\item For larger regression datasets, we use Sparse GP regression (\verb+gpflow.SGPR+, \citep{titsias2009variational}).
		\item For classification datasets, we use Variational Inference \citep{hensman2015scalable}. For datasets with ($N>200$), choose the number of inducing points $M = 200$, for SUSY and Churn modelling datasets, we choose  $M = 800$.
	\end{enumerate}
	We place a Gamma prior on the variance hyperparameters of the kernel, which are estimated using MAP. The lengthscales hyperparameters are estimated by maximum likelihood, or by maximising the ELBO, appropriately.  
	\item Construct the Sobol index for each component and each order according to equation (\ref{eq:sobol}), and construct a ranking. Truncate components when the (normalized) Sobol idex is below some threshold (default 0.01).
	\item Compute the posterior over the additive components identified in the above ranking, using equation (\ref{eq:posterior_component}).
	\item Predict for test points by summing over the identified components.
\end{enumerate}

\section{Two-dimensional Toy Example}
\label{sec:toy}
Additional experimental results for the two-dimensional example with (unconstrained) squared exponential kernel.

\begin{figure}[h!]
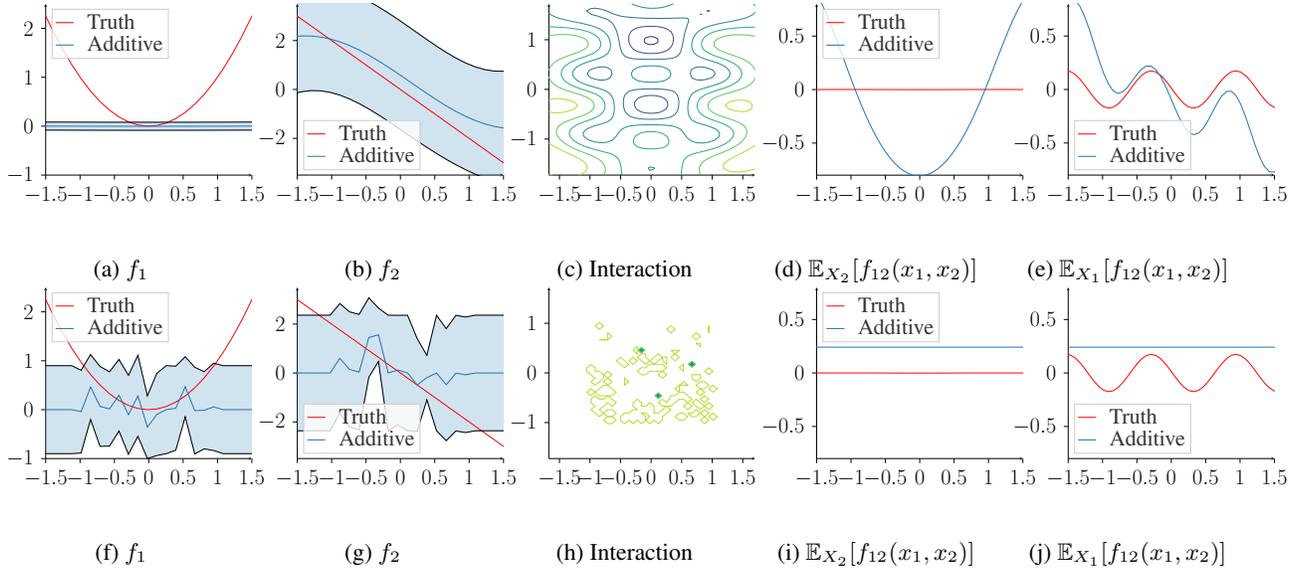

	\centering
	\begin{subfigure}{0.19\textwidth}
		\input{tikz/2d/2d_rbf_f1_3.tex}
		\caption{$f_1$}
		\label{fig:rbd_f1_1}
	\end{subfigure} 
	\begin{subfigure}{0.19\textwidth}
		\input{tikz/2d/2d_rbf_f2_3.tex}
		\caption{$f_2$}
		\label{fig:rbd_f2_1}
	\end{subfigure}
	\begin{subfigure}{0.19\textwidth}
		\input{tikz/2d/2d_interaction_rbf_3.tex}
		\caption{Interaction}
	\end{subfigure}	
	\begin{subfigure}{0.19\textwidth}
		\input{tikz/2d/2d_rbf_interaction_f1_3.tex}
		\caption{$\E_{X_2}[f_{12}(x_1,x_2)] $}
		\label{fig:inter_rbf_f1_1}
	\end{subfigure}	
	\begin{subfigure}{0.19\textwidth}
		\input{tikz/2d/2d_rbf_interaction_f2_3.tex}
		\caption{$\E_{X_1}[f_{12}(x_1,x_2)]$ }
		\label{fig:inter_rbf_f2_1}
	\end{subfigure}

	\begin{subfigure}{0.19\textwidth}
		\input{tikz/2d/2d_rbf_f1_6.tex}
		\caption{$f_1$}
		\label{fig:rbd_f1_2}
	\end{subfigure} 
	\begin{subfigure}{0.19\textwidth}
		\input{tikz/2d/2d_rbf_f2_6.tex}
		\caption{$f_2$}
		\label{fig:rbd_f2_2}
	\end{subfigure}
	\begin{subfigure}{0.19\textwidth}
		\input{tikz/2d/2d_interaction_rbf_6.tex}
		\caption{Interaction}
	\end{subfigure}	
	\begin{subfigure}{0.19\textwidth}
		\input{tikz/2d/2d_rbf_interaction_f1_6.tex}
		\caption{$\E_{X_2}[f_{12}(x_1,x_2)] $}
		\label{fig:inter_rbf_f1_2}
	\end{subfigure}	
	\begin{subfigure}{0.19\textwidth}
		\input{tikz/2d/2d_rbf_interaction_f2_6.tex}
		\caption{$\E_{X_1}[f_{12}(x_1,x_2)]$ }
		\label{fig:inter_rbf_f2_2}
	\end{subfigure}
	\caption{Two dimensional experimental results for the additive GP model in \citet{duvenaud2011additive} with squared exponential kernel for the remaining two local optima. The red lines represent the true function, blue shaded area represent $\pm2$  standard deviation. From left to right: posterior of $f_1$; posterior of $f_2$; posterior for the interaction term; marginal plot for $f_1$ in the interaction term ($\E_{X_2} [f_{12}(x_1,x_2)]$); marginal plot for $f_2$ in the interaction term ($\E_{X_1} [f_{12}(x_1,x_2)]$). Note how the quadratic shape in Figure \ref{fig:rbd_f1_1} and the linear trend in Figure \ref{fig:rbd_f2_1} are captured in the higher order terms Figure \ref{fig:inter_rbf_f1_1}  and Figure \ref{fig:inter_rbf_f2_1}. Vice Versa, first order terms may also absorb effect from the interaction, as Figure \ref{fig:rbd_f1_2}  and  Figure \ref{fig:rbd_f2_2} show.  }
\end{figure}

\newpage
\section{Baseline Experimental Results}
\label{sec:UCI}
\subsection{Baseline Dataset Details}
\label{sec:uci_data}
\begin{table*}[h!]
	\scriptsize
	\begin{center}
		\begin{tabular}{ |c|c|c|c|c|c|c|c|c|c|c| } 
			\hline
			Data & AutoMP  & Housing & Concrete  & Pumadyn & Breast &Pima & Sonar& Ionosphere& Liver & Heart\\
			\hline
			n & 392 & 506 &  1030 &8192 & 449 & 768 & 208 & 351 & 345 &297 \\ 
			D &  7 & 13 &  8 &8 & 9 & 8 & 60 & 32 & 6 &13
			\\ 
			\hline
		\end{tabular}
		\caption{Number of data and dimensionality of baseline datasets.}
		\label{table:uci_data}
	\end{center}
\end{table*}

\subsection{Total Number of Terms for Baseline Datasets}
\label{sec:number_of_terms}

\begin{table*}[h!]
	\scriptsize
	\begin{center}
		\begin{tabular}{ |c|c|c|c|c|c|c|c|c|c|c| } 
			\hline
			Data & AutoMP  & Housing & Concrete  & Pumadyn & Breast &Pima & Sonar& Ionosphere& Liver & Heart\\
			\hline
			number of terms & 127 & 8191 &  255 & 255 & 255 & 162 & 1830 & 41448 & 56 &1092 \\ 
			\hline
		\end{tabular}
		\caption{Total number of additive terms in baseline datasets.}
		\label{table:number_of_terms}
	\end{center}
\end{table*}

\subsection{Model Performance}
\begin{figure}[h!]
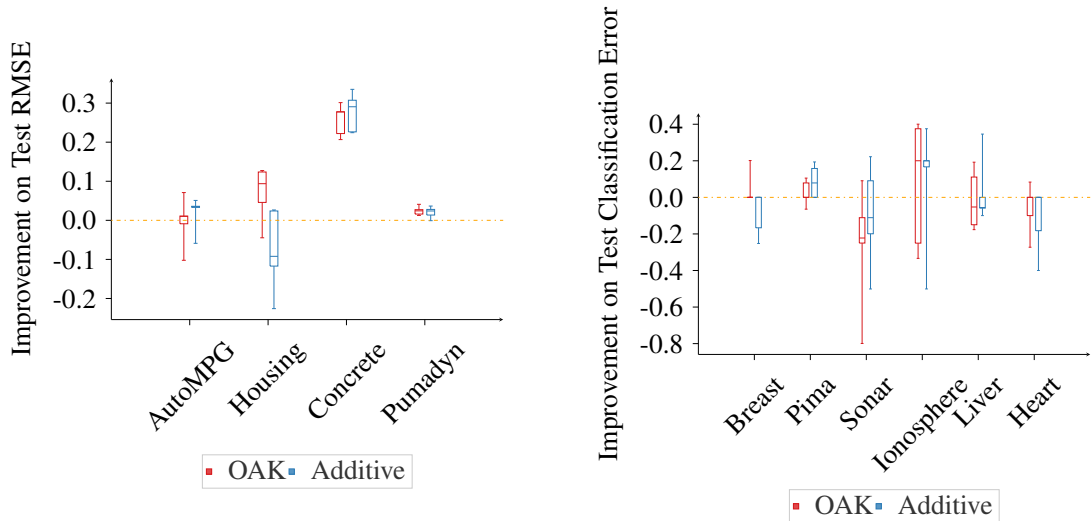

	\centering
	\begin{subfigure}{0.45\textwidth}
		\input{tikz/UCI/regression_error.tex}
	\end{subfigure} 
	\begin{subfigure}{0.45\textwidth}
		\input{tikz/UCI/classification_error.tex}
	\end{subfigure}
	\caption{Test RMSE relative improvement compared with GP with squared-exponential kernel for regression (left); and test classification percentage error for classification (right). Red and blue boxes represent mean and $\pm 1$ standard deviation over 5 train-test folds for the additive model and OAK model respectively. Horizontal axis represents different datasets; vertical axis represents model percentage improvement relative to the baseline model. Higher values are better.}
	\label{fig:uci_error}
\end{figure}

\begin{figure}[h!]
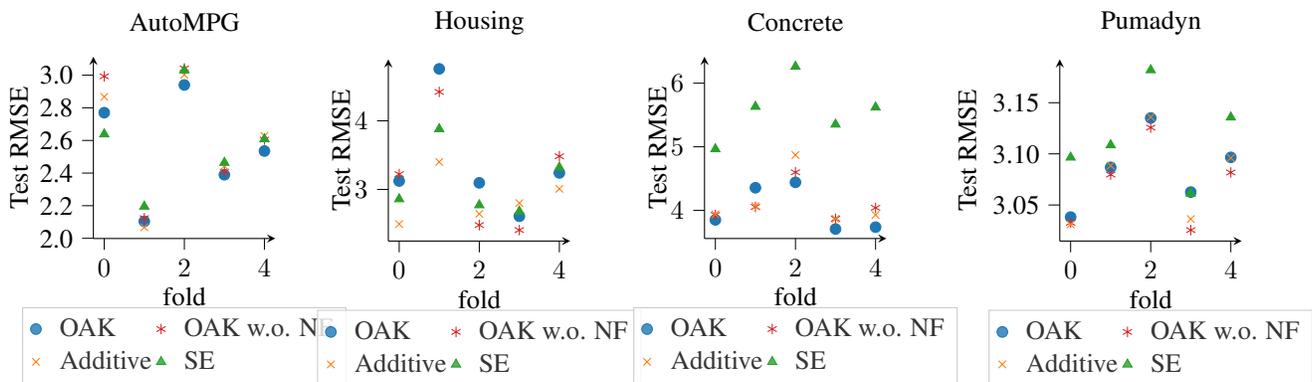

	\centering
	\begin{subfigure}{0.24\textwidth}
		\input{tikz/UCI/autoMPG_rmse_nnf.tex}
	\end{subfigure} 
	\begin{subfigure}{0.24\textwidth}
	\input{tikz/UCI/Housing_rmse_nnf.tex}
	\end{subfigure}
	\begin{subfigure}{0.24\textwidth}
		\input{tikz/UCI/concrete_rmse_nnf.tex}
	\end{subfigure}	
	\begin{subfigure}{0.24\textwidth}
			\input{tikz/UCI/pumadyn_rmse_nnf.tex}
	\end{subfigure}
\caption{Test RMSE on regression datasets over 5 folds, lower is better. OAK w.o. NF stands for the OAK model without normalizing flow.}
\label{fig:UCI_regression}
\end{figure}

\begin{figure}[h!]
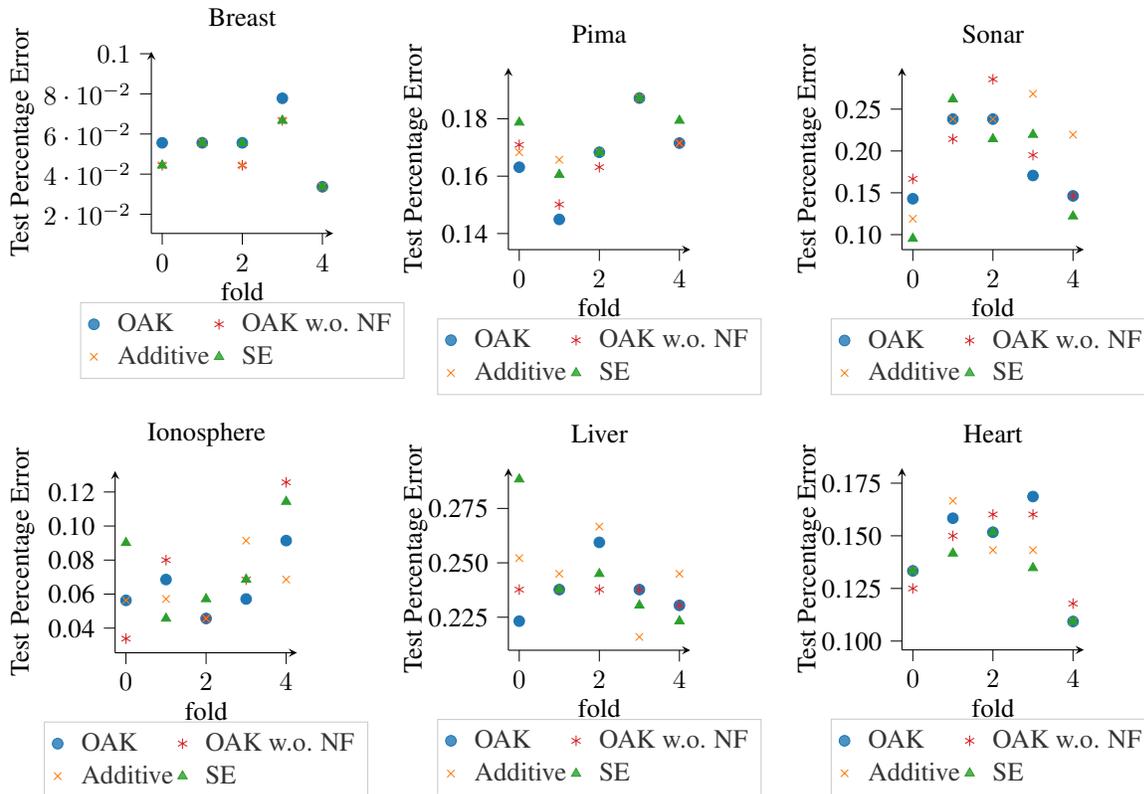

	\centering
	\begin{subfigure}{0.3\textwidth}
		\input{tikz/UCI/breast_percentage_error_nnf.tex}
	\end{subfigure} 
	\begin{subfigure}{0.3\textwidth}
		\input{tikz/UCI/pima_percentage_error_nnf.tex}
	\end{subfigure}
	\begin{subfigure}{0.3\textwidth}
		\input{tikz/UCI/sonar_percentage_error_nnf.tex}
	\end{subfigure}	

	\begin{subfigure}{0.3\textwidth}
		\input{tikz/UCI/ionosphere_percentage_error_nnf.tex}
	\end{subfigure}
		\begin{subfigure}{0.3\textwidth}
		\input{tikz/UCI/liver_percentage_error_nnf.tex}
	\end{subfigure}	
	\begin{subfigure}{0.3\textwidth}
		\input{tikz/UCI/heart_percentage_error_nnf.tex}
	\end{subfigure}
\caption{Test percenrage error on classification datasets over 5 folds, lower is better. OAK w.o. NF stands for the OAK model without normalizing flow.}
\label{fig:UCI_classification}
\end{figure}

\begin{figure}[h!]
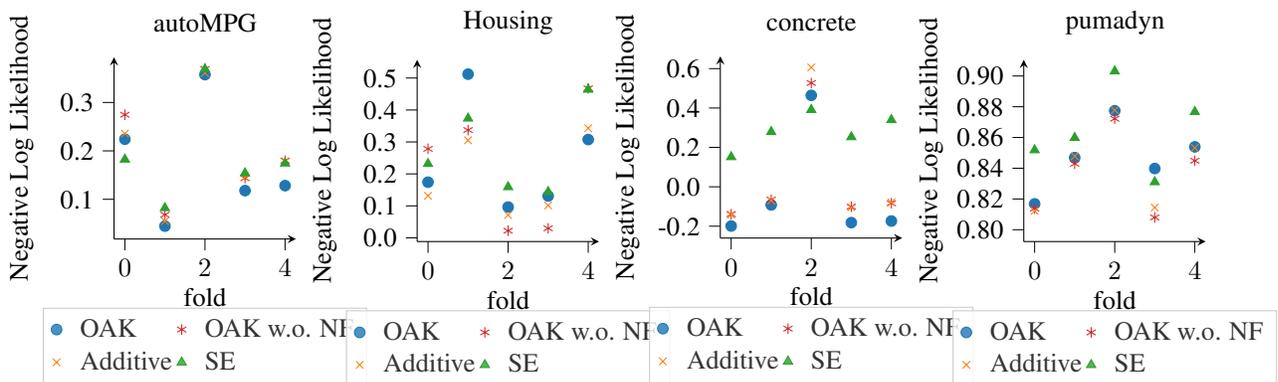

	\centering
	\begin{subfigure}{0.23\textwidth}
		\input{tikz/UCI/autoMPG_nll.tex}
	\end{subfigure} 
	\begin{subfigure}{0.23\textwidth}
		\input{tikz/UCI/Housing_nll.tex}
	\end{subfigure}
	\begin{subfigure}{0.23\textwidth}
		\input{tikz/UCI/concrete_nll.tex}
	\end{subfigure}	
	\begin{subfigure}{0.23\textwidth}
		\input{tikz/UCI/pumadyn_nll.tex}
	\end{subfigure}
	\caption{Negative log likelihood on regression datasets over 5 folds, lower is better. OAK w.o. NF stands for the OAK model without normalizing flow.}
	\label{fig:UCI_regression_nll}
\end{figure}

\begin{figure}[h!]
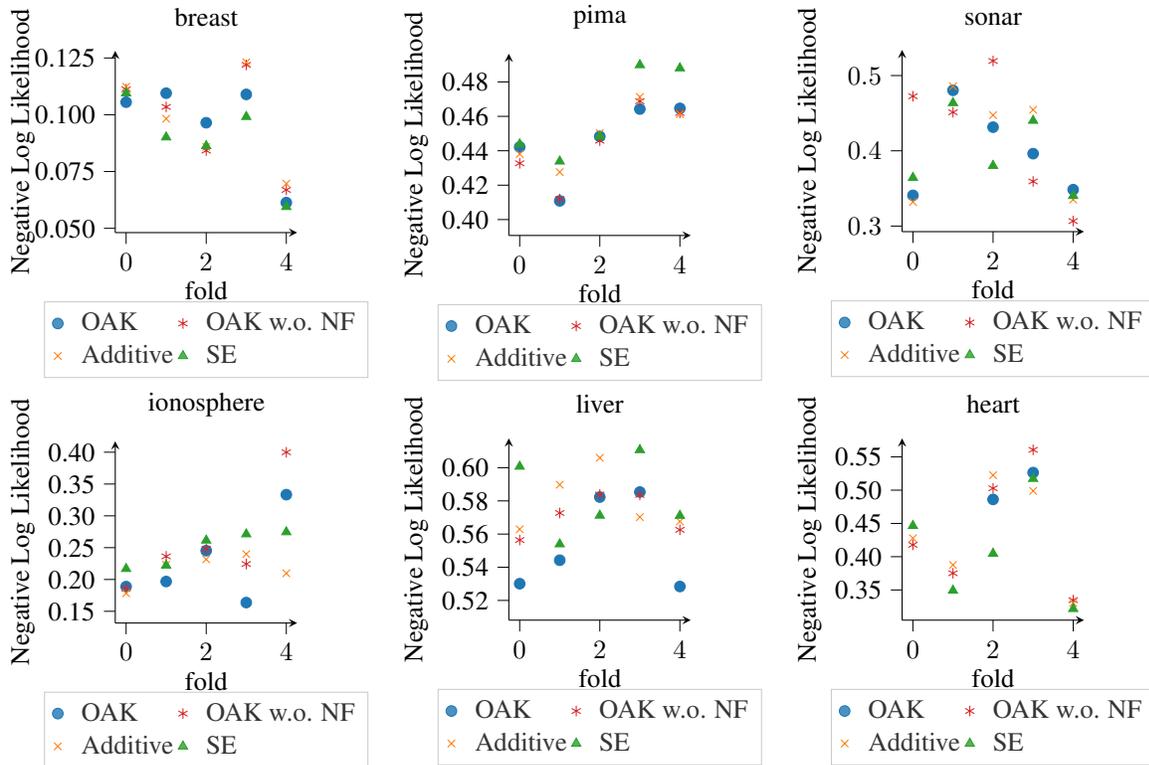

	\centering
	\begin{subfigure}{0.3\textwidth}
		\input{tikz/UCI/breast_nll.tex}
	\end{subfigure} 
	\begin{subfigure}{0.3\textwidth}
		\input{tikz/UCI/pima_nll.tex}
	\end{subfigure}
	\begin{subfigure}{0.3\textwidth}
		\input{tikz/UCI/sonar_nll.tex}
	\end{subfigure}	
	
	\begin{subfigure}{0.3\textwidth}
		\input{tikz/UCI/ionosphere_nll.tex}
	\end{subfigure}
	\begin{subfigure}{0.3\textwidth}
		\input{tikz/UCI/liver_nll.tex}
	\end{subfigure}	
	\begin{subfigure}{0.3\textwidth}
		\input{tikz/UCI/heart_nll.tex}
	\end{subfigure}
	\caption{Negative log likelihood on classification datasets over 5 folds, lower is better. OAK w.o. NF stands for the OAK model without normalizing flow.}
	\label{fig:UCI_classification_nll}
\end{figure}

Model performance for baseline dataset experiments are displayed in Figure \ref{fig:uci_error}, where we compare percentage improvement relative to the baseline model (full GP with squared exponential kernel) for our constrained kernel and the non-constrained counterparts in \citet{duvenaud2011additive}. Positive values indicate superior performance compared with the baseline. Detailed performance on each of the train-split fold can be found in Figure \ref{fig:UCI_regression}, \ref{fig:UCI_classification}, \ref{fig:UCI_regression_nll} and \ref{fig:UCI_classification_nll}.

\subsection{Order Variance Hyperparameter Comparison}
\label{sec:relative_var}
\begin{figure}[h!]
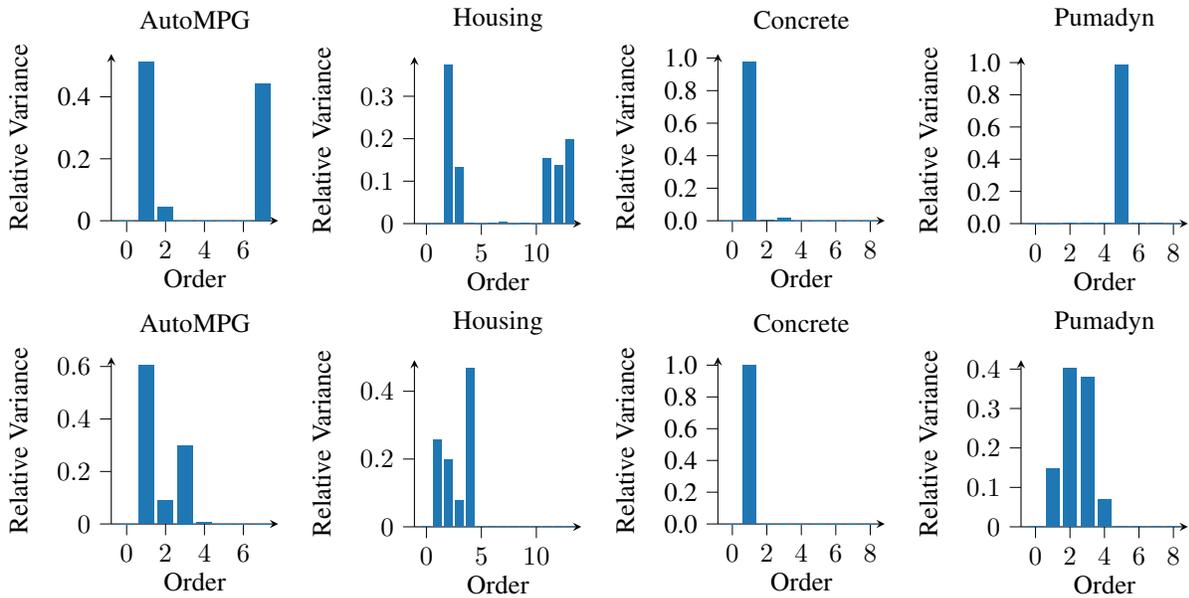

	\centering
	\vspace*{-3mm}
	\begin{subfigure}{0.23\textwidth}
		\input{tikz/UCI/autoMPG_relative_var_duvenaud.tex}
	\end{subfigure} 
	\begin{subfigure}{0.23\textwidth}
		\input{tikz/UCI/Housing_relative_var_duvenaud.tex}
	\end{subfigure} 
	\begin{subfigure}{0.23\textwidth}
		\input{tikz/UCI/concrete_relative_var_duvenaud.tex}
	\end{subfigure} 
	\begin{subfigure}{0.23\textwidth}
		\input{tikz/UCI/pumadyn_relative_var_duvenaud.tex}
	\end{subfigure}

	\begin{subfigure}{0.23\textwidth}
		\input{tikz/UCI/autoMPG_relative_var_oak.tex}
	\end{subfigure} 
	\begin{subfigure}{0.23\textwidth}
		\input{tikz/UCI/Housing_relative_var_oak.tex}
	\end{subfigure} 
	\begin{subfigure}{0.23\textwidth}
		\input{tikz/UCI/concrete_relative_var_oak.tex}
	\end{subfigure} 
	\begin{subfigure}{0.23\textwidth}
		\input{tikz/UCI/pumadyn_relative_var_oak.tex}
	\end{subfigure} 
	\caption{Normalized order variance hyperparameter on the UCI regression datasets. Top: kernel used in \citet{duvenaud2011additive}; bottom: OAK model. OAK requires lower dimensional orders of interactions with similar performance. Results are averaged over 5 folds.}
	\label{fig:relative_var_regression}
\end{figure}

\begin{figure}[h!]
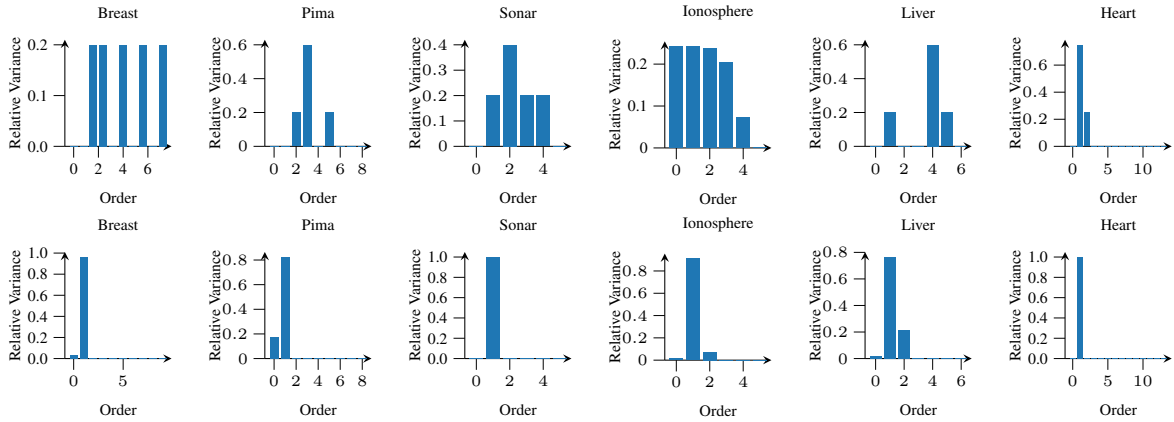

	\centering
	\vspace*{-3mm}
	\begin{subfigure}{0.15\textwidth}
		\input{tikz/UCI/breast_relative_var_duvenaud.tex}
	\end{subfigure} 
	\begin{subfigure}{0.15\textwidth}
		\input{tikz/UCI/pima_relative_var_duvenaud.tex}
	\end{subfigure} 
	\begin{subfigure}{0.15\textwidth}
		\input{tikz/UCI/sonar_relative_var_duvenaud.tex}
	\end{subfigure} 
	\begin{subfigure}{0.15\textwidth}
		\input{tikz/UCI/ionosphere_relative_var_duvenaud.tex}
	\end{subfigure} 
	\begin{subfigure}{0.15\textwidth}
	\input{tikz/UCI/liver_relative_var_duvenaud.tex}
\end{subfigure} 
\begin{subfigure}{0.15\textwidth}
	\input{tikz/UCI/heart_relative_var_duvenaud.tex}
\end{subfigure} 	
	
	\begin{subfigure}{0.15\textwidth}
		\input{tikz/UCI/breast_relative_var_oak.tex}
	\end{subfigure} 
	\begin{subfigure}{0.15\textwidth}
		\input{tikz/UCI/pima_relative_var_oak.tex}
	\end{subfigure} 
	\begin{subfigure}{0.15\textwidth}
		\input{tikz/UCI/sonar_relative_var_oak.tex}
	\end{subfigure} 
	\begin{subfigure}{0.15\textwidth}
		\input{tikz/UCI/ionosphere_relative_var_oak.tex}
	\end{subfigure} 
	\begin{subfigure}{0.15\textwidth}
	\input{tikz/UCI/liver_relative_var_oak.tex}
\end{subfigure} 
\begin{subfigure}{0.15\textwidth}
	\input{tikz/UCI/heart_relative_var_oak.tex}
\end{subfigure} 
	\caption{Normalized order variance hyperparameter on the UCI classification datasets. Top: kernel used in \citet{duvenaud2011additive}; bottom: OAK model. OAK requires lower dimensional orders of interactions with similar performance. We have truncated the maximum order of interaction to 4 for Sonar and Ionosphere datasets. Results are averaged over 5 folds.}
	\label{fig:relative_var_classification}
\end{figure}

We compare the normalized variance hyperparameter $\frac{\sigma_d^2}{\sum_{d=1}^D \sigma_d^2}$ of each order $d$ of interaction between OAK and \citet{duvenaud2011additive}, as shown in Figure \ref{fig:relative_var_regression} and \ref{fig:relative_var_classification}. The results further verify that OAK model is more parsimonious and requires lower order interactions for all datasets.

\subsection{Normalizing Flow Ablation Study}
\label{sec:nf}
\begin{figure*}[h!]
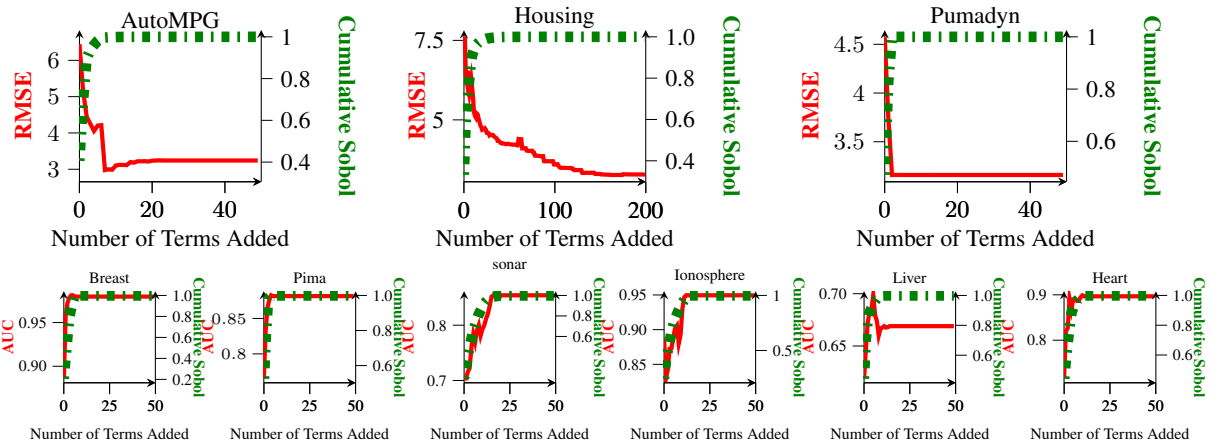

	\centering
	\begin{subfigure}[b]{0.3\textwidth}
		\input{tikz/UCI/autoMPG_cut_metric_and_sobol_nnf.tex}
	\end{subfigure} 
	\begin{subfigure}[b]{0.3\textwidth}
		\input{tikz/UCI/Housing_cut_metric_and_sobol_nnf.tex}
	\end{subfigure} 
	\begin{subfigure}[b]{0.3\textwidth}
		\input{tikz/UCI/pumadyn_cut_metric_and_sobol_nnf.tex}
	\end{subfigure} 

	\begin{subfigure}[b]{0.15\textwidth}
	\input{tikz/UCI/breast_cut_metric_and_sobol_nnf.tex}
	\end{subfigure} 
	\begin{subfigure}[b]{0.15\textwidth}
	\input{tikz/UCI/pima_cut_metric_and_sobol_nnf.tex}
	\end{subfigure} 
	\begin{subfigure}[b]{0.15\textwidth}
	\input{tikz/UCI/sonar_cut_metric_and_sobol_nnf.tex}
	\end{subfigure} 
	\begin{subfigure}[b]{0.15\textwidth}
	\input{tikz/UCI/ionosphere_cut_metric_and_sobol_nnf.tex}
	\end{subfigure} 
	\begin{subfigure}[b]{0.15\textwidth}
	\input{tikz/UCI/liver_cut_metric_and_sobol_nnf.tex}
	\end{subfigure} 
	\begin{subfigure}[b]{0.15\textwidth}
	\input{tikz/UCI/heart_cut_metric_and_sobol_nnf.tex}
	\end{subfigure} 
	\caption{Model performance and cumulative Sobol index versus number of terms added ranked by the Sobol index, 
	without normalizing flow. For regression problems (top), 
	we use test RMSE as the evaluation metric. 
	Note that we did not include result for the Concrete dataset because the NF was not sufficient to transform the data and 
	we used the empirical measure for it in Figure \ref{fig:sobol_plots}: in this case the predictive performance was not affected, but the parsimony of the result (i.e. the number of terms needed to reach the same performance) was. 
	For classification problems (bottom), we use test area-under-the-curve (AuC) metric. 
	Red solid lines represent test RMSE (top) and test AuC (bottom), 
	green dashed lines represent cumulative (normalized) Sobol index.}
	\label{fig:sobol_plots_nnf}
\end{figure*}

Normalizing flow plays a role similar to data centering: we transform each continuous feature to be closer to Gaussian. The bijective function in the flow
is a composition of shifting, scaling and sinharcsinh transformation. The parameters of the bijective functions are learned and fixed before fitting the GP model, using only the input data, not in conjunction with the hyperparametrs. For non-continuous input features we do not apply any transformation, but use the orthogonal discrete kernel described in Appendix \ref{sec:coregional}. We have performed an ablation study and ran experiments on all the baseline datasets where we standardize the inputs instead of using the flow. The model performance is similar (see Figure \ref{fig:UCI_regression}, \ref{fig:UCI_classification}, \ref{fig:UCI_regression_nll} and \ref{fig:UCI_classification_nll}) but the resulting model tends to be less parsimonious, especially for the Housing dataset, see Figure \ref{fig:sobol_plots_nnf} for details. 

\newpage
\subsection{Comparison with Kernel in \citet{duvenaud2011additive}}
\label{sec:product_kernel}
\begin{figure}[h!]
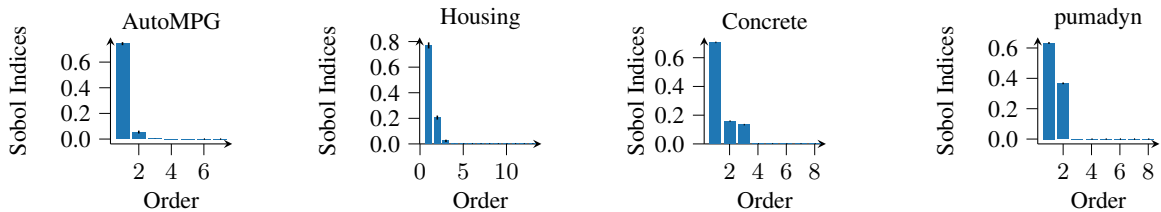

	\small
	\begin{center}
			\begin{subfigure}[b]{0.235\textwidth}
				\input{tikz/UCI/autoMPG_durrande_prod_sobol_order.tex}
			\end{subfigure} 
			\begin{subfigure}[b]{0.235\textwidth}
				\input{tikz/UCI/housing_durrande_prod_sobol_order.tex}
			\end{subfigure} 
			\begin{subfigure}[b]{0.235\textwidth}
				\input{tikz/UCI/concrete_durrande_prod_sobol_order.tex}
			\end{subfigure} 
			\begin{subfigure}[b]{0.235\textwidth}
				\input{tikz/UCI/pumadyn_durrande_prod_sobol_order.tex}
			\end{subfigure} 
	\end{center}
	\caption{Cumulative Sobol Indices with kernel of the form $\prod_d (1 + \tilde{k}_d)$  in \citet{duvenaud2011additive} using constrained kernel. The model performance is similar to OAK but the resulting model tends to be less parsimonious: e.g., Concrete dataset needs 3rd order terms with normalized Sobol indices = 0.71, 0.16, 0.13 for 1st, 2nd and 3rd order respectively as opposed to 0.971, 0.026, 0.003 with OAK in Figure \ref{fig:sobol}. }
	\label{fig:product_kernel}
\end{figure}

The kernel $\prod_d (1 + \tilde{k}_d)$ restricts the lengthscales and variances of the kernels to be the same
for lower and higher order terms, e.g., if two features are important in their main effect, the interaction between them
 will also be important, which may result in a less parsimonious model as higher order terms cannot be downweighted during inference. We have
conducted experiments using this kernel for comparison, with results shown in Figure \ref{fig:product_kernel}. We found the kernel is harder to optimize and numerically unstable, the model performance is similar but the resulting model is less parsimonious: e.g., Concrete dataset
needs 3rd order terms (with normalized Sobol indices = 0.71, 0.16, 0.13 for 1st, 2nd and 3rd order respectively, as opposed to 0.971, 0.026, 0.003 with OAK in Figure \ref{fig:sobol}. 

\subsection{Number of Inducing Points Needed}
\label{sec:pumadyn_rmse}
When kernels are combined through a product, the eigenspectrum is the outer-product of the spectra of the components (Corollary 3 in \citet{burt2019rates}). This is what leads to the exponential scaling of the
number of inducing points with the dimension of the problem, $M = \mathcal{O}(log^D N)$. When we add kernels together, the
eigenspectrum is simply the concatenation of the spectrum of each component, so the resulting scaling is linear.

\begin{figure}
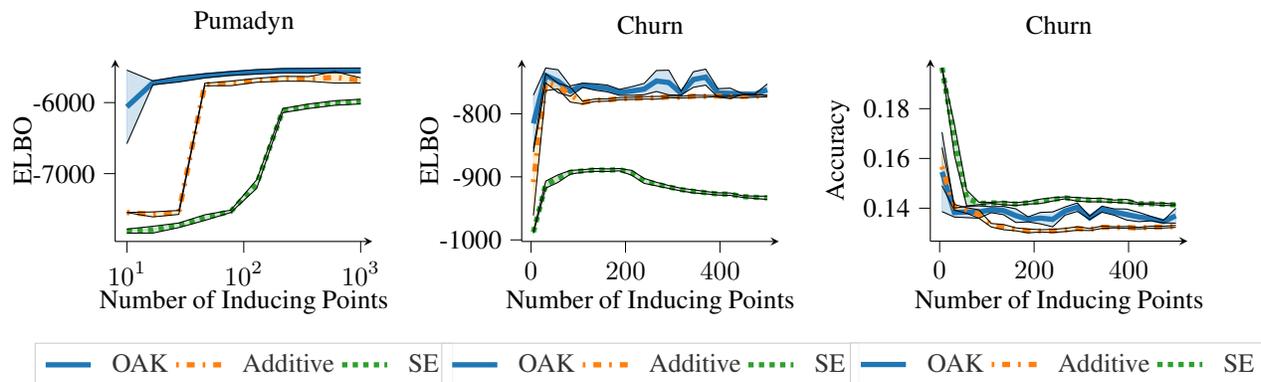

	\centering
		\begin{subfigure}[b]{0.31\textwidth}
			\input{tikz/UCI/pumadyn-inducing-elbo.tex}
			\label{fig:inducing_pumadyn}
	\end{subfigure} 
	\begin{subfigure}[b]{0.31\textwidth}
		\input{tikz/UCI/churn-inducing-elbo.tex}
		\label{fig:elbo_churn}
	\end{subfigure} 
	\begin{subfigure}[b]{0.31\textwidth}
			\input{tikz/UCI/churn-inducing-accuracy.tex}
			\label{fig:acc_churn}
	\end{subfigure} 
	\caption{Model performance versus varying number of inducing points on the test set. Results are averaged over 5 repetitions for the Pumadyn dataset and 10 repetitions for the Churn dataset. Shaded area represents $\pm 1$ standard deviation. Note that test ELBO is not always monotone on the Churn data, we attribute this to the difficulty of finding local optima. }
	\label{fig:pumadyn_rmse}
\end{figure}

Additional experiments on number of inducing points needed for the pumadyn and Churn datasets can be found in Figure \ref{fig:pumadyn_rmse}. The number of inducing points needed for OAK is smaller than that for the non-orthogonal model and the full GP model. Note that although the ELBO values are not directly comparable due to the normalizing flow used for some of the models, we can observe that the OAK model converges much faster than its counterparts.

\section{Additional Benchmark Experiments}
The evaluation results for the entire set of datasets\footnote{Data and code for other methods are taken from \url{https://github.com/hughsalimbeni/bayesian_benchmarks.} } summarized in Table \ref{table:additional_benchmarks} 
can be found in Tables \ref{table:regree_rmse}-\ref{table:class_ll}. 
Values outside $[-1000,1000]$ are denoted as NaN. Results are averaged over 10 train-test splits, values in brackets represent one standard deviation.
\label{sec:additional_benchmarks}
\begin{table}[h!]
	\tiny
	\centering
	\begin{tabular}{lllllllllll}
  \toprule
     dataset &     N &   D &          OAK &              Linear &         SVGP &          SVM &          KNN &          GBM &           AdaBoost &          MLP \\
  \midrule
      boston &   506 &  13 & 0.290(0.036) &     0.444(0.044) & 0.312(0.030) & 0.267(0.037) & 0.380(0.059) & 0.282(0.020) & 0.349(0.026) & 0.299(0.030) \\
      energy &   768 &   8 & 0.036(0.010) &     0.300(0.034) & 0.048(0.005) & 0.227(0.027) & 0.218(0.029) & 0.047(0.005) & 0.191(0.006) & 0.193(0.020) \\
       naval & 11934 &  14 & 0.164(0.313) &     0.394(0.007) & 0.004(0.001) & 0.215(0.006) & 0.104(0.006) & 0.263(0.007) & 0.885(0.017) & 0.051(0.007) \\
       power &  9568 &   4 & 0.234(0.009) &     0.267(0.008) & 0.237(0.009) & 0.234(0.009) & 0.219(0.008) & 0.226(0.008) & 0.327(0.012) & 0.236(0.009) \\
     winered &  1599 &  11 & 0.775(0.044) &     0.808(0.046) & 0.926(0.145) & 0.768(0.055) & 0.825(0.063) & 0.762(0.046) & 0.774(0.054) & 0.773(0.055) \\
   winewhite &  4898 &  11 & 0.827(0.079) &     0.847(0.033) & 0.837(0.084) & 0.768(0.021) & 0.788(0.020) & 0.768(0.020) & 0.826(0.022) & 0.763(0.023) \\
     protein & 45730 &   9 & 0.987(0.032) &     0.850(0.004) & 0.782(0.007) & 0.764(0.008) & 0.623(0.007) & 0.768(0.007) & 0.933(0.012) & 0.707(0.022) \\
       yacht &   308 &   6 & 0.032(0.012) &     0.608(0.048) & 0.048(0.016) & 0.419(0.092) & 0.668(0.144) & 0.044(0.014) & 0.103(0.023) & 0.244(0.051) \\
     airfoil &  1503 &   5 & 0.837(0.174) &     0.721(0.047) & 0.456(0.033) & 0.486(0.038) & 0.429(0.035) & 0.387(0.043) & 0.573(0.029) & 0.412(0.041) \\
      forest &   517 &  12 & 1.030(0.100) &     1.018(0.106) & 0.995(0.025) & 1.100(0.139) & 1.117(0.142) & 1.069(0.131) & 1.092(0.093) & 1.077(0.115) \\
  parkinsons &   195 &  23 & 0.373(0.140) &     0.871(0.021) & 0.635(0.021) & 0.544(0.022) & 0.384(0.024) & 0.245(0.008) & 0.587(0.020) & 0.283(0.017) \\
       stock &   536 &  11 & 0.305(0.049) &     0.286(0.025) & 0.286(0.027) & 0.465(0.136) & 0.579(0.094) & 0.348(0.058) & 0.363(0.071) & 0.308(0.025) \\
   fertility &   100 &  10 & 0.799(0.192) &     0.900(0.229) & 0.975(0.295) & 0.975(0.250) & 1.055(0.287) & 1.032(0.225) & 0.904(0.209) & 1.020(0.233) \\
     machine &   209 &   7 & 0.281(0.044) &     0.435(0.053) & 0.398(0.048) & 0.419(0.054) & 0.417(0.076) & 0.338(0.043) & 0.368(0.037) & 0.393(0.044) \\
    pendulum &   630 &   9 & 0.443(0.099) &     0.862(0.164) & 0.653(0.136) & 0.654(0.188) & 0.626(0.132) & 0.772(0.110) & 0.810(0.134) & 0.659(0.140) \\
       servo &   167 &   4 & 0.312(0.069) &     0.607(0.068) & 0.299(0.074) & 0.343(0.060) & 0.454(0.070) & 0.270(0.070) & 0.383(0.062) & 0.364(0.060) \\
        wine &   178 &  14 & 0.449(0.033) &     0.564(0.029) & 0.469(0.034) & 0.440(0.041) & 0.562(0.045) & 0.461(0.031) & 0.620(0.041) & 0.436(0.038) \\
  tamielectr & 45781 &   3 & 1.001(0.005) &     1.001(0.005) & 1.001(0.005) & 1.002(0.005) & 1.099(0.007) & 1.002(0.005) & 1.002(0.005) & 1.002(0.005) \\
      kin40k & 40000 &   8 & 0.581(0.019) &     1.000(0.013) & 0.682(0.016) & 0.205(0.004) & 0.392(0.005) & 0.842(0.010) & 0.939(0.013) & 0.187(0.007) \\
         gas &  2565 & 128 & 0.254(0.078) & 112.965(333.613) & 0.182(0.041) & 0.227(0.101) & 0.119(0.037) & 0.117(0.029) & 0.313(0.025) & 0.496(0.595) \\
  keggdirect & 48827 &  20 & 0.129(0.065) &              nan & 0.109(0.005) & 0.102(0.002) & 0.097(0.004) & 0.094(0.003) & 0.201(0.003) & 0.199(0.318) \\
        bike & 17379 &  17 & 0.023(0.008) &     0.517(0.008) & 0.353(0.006) & 0.262(0.008) & 0.454(0.011) & 0.020(0.001) & 0.124(0.004) & 0.065(0.008) \\
         pol & 15000 &  26 & 0.848(0.131) &     0.736(0.011) & 0.396(0.010) & 0.335(0.006) & 0.215(0.013) & 0.256(0.008) & 0.492(0.017) & 0.151(0.007) \\
   elevators & 16599 &  18 & 0.379(0.007) &   14.600(21.603) & 0.394(0.007) & 0.392(0.007) & 0.602(0.016) & 0.502(0.014) & 0.776(0.014) & 0.359(0.013) \\
         avg &       &     &        0.475 &            6.157 &        0.478 &        0.484 &        0.518 &        0.455 &        0.581 &        0.445 \\
      median &       &     &        0.376 &            0.736 &        0.397 &        0.419 &        0.454 &        0.343 &        0.580 &        0.361 \\
    avg rank &       &     &        3.583 &            6.625 &        4.083 &        4.208 &        4.958 &        3.208 &        5.750 &        3.583 \\
  \bottomrule
  \end{tabular}
  	
	\caption{Test RMSE for regression tasks on additional benchmark datasets, lower is better.}
	\label{table:regree_rmse}
\end{table}

\begin{table}[h!]
	\tiny
	\centering
	\begin{tabular}{lllllllllll}
  \toprule
     dataset &     N &   D &           OAK &              Linear &         SVGP &          SVM &          KNN &          GBM &           AdaBoost &          MLP \\
  \midrule
      boston &   506 &  13 & -0.122(0.157) & -0.644(0.066) & -0.281(0.058) & -0.157(0.083) & -0.467(0.134) & -0.637(0.250) & -0.388(0.095) &  -0.248(0.140) \\
      energy &   768 &   8 &  1.923(0.308) & -0.220(0.114) &  1.609(0.081) &  0.038(0.159) & -0.021(0.254) &  1.603(0.154) &  0.235(0.035) &   0.194(0.117) \\
       naval & 11934 &  14 &  1.932(1.525) & -0.489(0.017) &  3.957(0.133) &  0.120(0.028) &  0.740(0.109) & -0.088(0.030) & -1.297(0.019) &   1.561(0.144) \\
       power &  9568 &   4 &  0.030(0.037) & -0.098(0.031) &  0.018(0.036) &  0.034(0.038) &  0.046(0.056) &  0.066(0.042) & -0.304(0.037) &   0.025(0.037) \\
     winered &  1599 &  11 & -1.166(0.059) & -1.208(0.060) & -1.507(0.517) & -1.174(0.095) & -1.280(0.121) & -1.206(0.099) & -1.170(0.081) &  -1.204(0.109) \\
   winewhite &  4898 &  11 & -1.224(0.091) & -1.254(0.039) & -1.236(0.095) & -1.161(0.031) & -1.230(0.040) & -1.161(0.031) & -1.229(0.028) &  -1.160(0.038) \\
     protein & 45730 &   9 & -1.407(0.030) & -1.257(0.005) & -1.172(0.008) & -1.150(0.011) & -1.013(0.018) & -1.156(0.009) & -1.350(0.013) &  -1.073(0.031) \\
       yacht &   308 &   6 &  1.320(1.503) & -0.929(0.083) &  1.715(0.237) & -0.614(0.287) & -1.152(0.329) & -0.597(2.242) &  0.799(0.351) &  -0.090(0.318) \\
     airfoil &  1503 &   5 & -1.395(0.600) & -1.096(0.070) & -0.650(0.072) & -0.711(0.093) & -0.693(0.149) & -0.496(0.139) & -0.865(0.054) &  -0.548(0.119) \\
      forest &   517 &  12 & -1.473(0.119) & -1.447(0.121) & -1.893(0.503) & -1.582(0.206) & -1.600(0.204) & -1.753(0.321) & -1.557(0.129) &  -1.594(0.196) \\
  parkinsons &   195 &  23 & -0.415(0.419) & -1.282(0.025) & -0.976(0.026) & -0.813(0.045) & -0.555(0.111) & -0.012(0.035) & -0.886(0.034) &  -0.243(0.107) \\
       stock &   536 &  11 & -0.199(0.111) & -0.175(0.079) & -0.173(0.078) & -1.090(0.975) & -0.975(0.287) & -1.100(0.687) & -0.486(0.331) &  -0.344(0.164) \\
   fertility &   100 &  10 & -1.244(0.239) & -1.376(0.362) & -1.461(0.425) & -1.676(0.735) & -1.631(0.538) & -3.890(1.808) & -1.608(0.608) &  -2.800(1.437) \\
     machine &   209 &   7 & -0.162(0.153) & -0.598(0.134) & -0.519(0.132) & -0.603(0.190) & -0.629(0.279) & -1.566(0.731) & -0.507(0.174) &  -0.510(0.145) \\
    pendulum &   630 &   9 &  0.309(0.978) & -1.299(0.209) & -0.912(0.184) & -1.129(0.462) & -1.020(0.308) & -2.375(0.766) & -1.329(0.317) &  -1.354(0.585) \\
       servo &   167 &   4 & -0.402(0.522) & -0.929(0.098) & -0.265(0.211) & -0.400(0.245) & -0.783(0.304) & -0.418(0.660) & -0.513(0.225) &  -0.432(0.207) \\
        wine &   178 &  14 & -0.613(0.068) & -0.849(0.054) & -0.660(0.070) & -0.624(0.127) & -0.900(0.129) & -0.706(0.109) & -0.947(0.075) &  -0.651(0.143) \\
  tamielectr & 45781 &   3 & -1.461(0.117) & -1.420(0.005) & -1.420(0.005) & -1.421(0.005) & -1.561(0.010) & -1.422(0.005) & -1.420(0.005) &  -1.421(0.005) \\
      kin40k & 40000 &   8 & -0.874(0.030) & -1.419(0.013) & -1.034(0.022) &  0.164(0.022) & -0.529(0.019) & -1.247(0.012) & -1.357(0.015) &   0.253(0.037) \\
         gas &  2565 & 128 & -0.166(0.260) &           nan &  0.292(0.103) & -0.051(0.379) &  0.646(0.357) & -0.101(0.894) & -0.275(0.098) &  -3.604(9.953) \\
  keggdirect & 48827 &  20 &  0.591(0.526) &           nan &  0.853(0.027) &  0.853(0.030) &  0.897(0.060) &  0.945(0.033) &  0.184(0.015) & -7.418(25.044) \\
        bike & 17379 &  17 &  2.416(0.383) & -0.759(0.015) & -0.379(0.017) & -0.085(0.033) & -0.688(0.040) &  2.468(0.050) &  0.666(0.037) &   1.315(0.129) \\
         pol & 15000 &  26 & -1.246(0.158) & -1.112(0.015) & -0.506(0.024) & -0.327(0.018) &  0.052(0.095) & -0.058(0.033) & -0.710(0.033) &   0.348(0.073) \\
   elevators & 16599 &  18 & -0.450(0.023) &           nan & -0.488(0.015) & -0.489(0.022) & -0.975(0.044) & -0.733(0.030) & -1.196(0.021) &  -0.397(0.039) \\
         avg &       &     &        -0.229 &        -0.946 &        -0.295 &        -0.585 &        -0.638 &        -0.652 &        -0.730 &         -0.891 \\
      median &       &     &        -0.409 &        -1.096 &        -0.512 &        -0.609 &        -0.738 &        -0.671 &        -0.875 &         -0.471 \\
    avg rank &       &     &         5.583 &         3.625 &         5.042 &         4.833 &         3.917 &         4.292 &         3.583 &          5.125 \\
  \bottomrule
  \end{tabular}	
	\caption{Test log likelihood for regression tasks on additional benchmark datasets, higher is better.}
	\label{table:regree_ll}
\end{table}

\begin{table}[h!]
	\tiny
	\centering
	\begin{tabular}{lllllllllll}
   \toprule
      dataset &    N &   D &          OAK &              Linear &         SVGP &          SVM &          KNN &          GBM &           AdaBoost &          MLP \\
   \midrule
   acute-infl &  120 &   7 & 1.000(0.000) & 1.000(0.000) & 1.000(0.000) & 1.000(0.000) & 1.000(0.000) & 1.000(0.000) & 0.958(0.072) & 1.000(0.000) \\
   acute-neph &  120 &   7 & 1.000(0.000) & 1.000(0.000) & 1.000(0.000) & 1.000(0.000) & 1.000(0.000) & 1.000(0.000) & 0.992(0.025) & 1.000(0.000) \\
         bank & 4521 &  17 & 0.898(0.016) & 0.891(0.018) & 0.890(0.017) & 0.891(0.013) & 0.890(0.014) & 0.900(0.011) & 0.892(0.015) & 0.893(0.012) \\
        blood &  748 &   5 & 0.740(0.007) & 0.780(0.055) & 0.787(0.051) & 0.781(0.050) & 0.767(0.041) & 0.768(0.041) & 0.776(0.046) & 0.793(0.043) \\
   chess-krvk & 3196 &  37 & 0.960(0.021) & 0.980(0.005) & 0.980(0.006) & 0.993(0.004) & 0.959(0.007) & 0.999(0.001) & 0.999(0.001) & 0.997(0.003) \\
   congressio &  435 &  17 & 0.616(0.050) & 0.616(0.042) & 0.605(0.050) & 0.630(0.061) & 0.568(0.067) & 0.584(0.070) & 0.582(0.056) & 0.566(0.059) \\
   conn-bench &  208 &  61 & 0.990(0.019) & 0.986(0.030) & 0.976(0.038) & 0.971(0.032) & 0.900(0.054) & 1.000(0.000) & 1.000(0.000) & 0.929(0.053) \\
   credit-app &  690 &  16 & 0.888(0.067) & 0.849(0.051) & 0.851(0.045) & 0.833(0.036) & 0.830(0.044) & 0.967(0.025) & 0.971(0.016) & 0.858(0.037) \\
   cylinder-b &  512 &  36 & 0.752(0.060) & 0.727(0.042) & 0.735(0.034) & 0.779(0.031) & 0.785(0.053) & 0.810(0.045) & 0.738(0.030) & 0.767(0.029) \\
   echocardio &  131 &  11 & 0.879(0.091) & 0.850(0.126) & 0.864(0.117) & 0.850(0.098) & 0.814(0.136) & 0.843(0.070) & 0.843(0.083) & 0.843(0.114) \\
    fertility &  100 &  10 & 0.900(0.050) & 0.900(0.063) & 0.920(0.060) & 0.920(0.060) & 0.910(0.054) & 0.860(0.092) & 0.870(0.078) & 0.890(0.070) \\
   haberman-s &  306 &   4 & 0.758(0.089) & 0.755(0.087) & 0.765(0.089) & 0.745(0.065) & 0.694(0.070) & 0.713(0.101) & 0.745(0.087) & 0.745(0.070) \\
   heart-hung &  294 &  13 & 1.000(0.000) & 0.997(0.010) & 0.997(0.010) & 0.970(0.023) & 0.863(0.055) & 1.000(0.000) & 1.000(0.000) & 0.990(0.015) \\
    hepatitis &  155 &  20 & 0.819(0.071) & 0.794(0.097) & 0.856(0.056) & 0.844(0.075) & 0.819(0.076) & 0.812(0.079) & 0.787(0.098) & 0.844(0.075) \\
   hill-valle & 1212 & 101 & 0.483(0.048) & 0.556(0.043) & 0.484(0.043) & 0.493(0.040) & 0.507(0.031) & 0.520(0.036) & 0.517(0.038) & 0.526(0.061) \\
   horse-coli &  368 &  26 & 0.824(0.039) & 0.832(0.055) & 0.824(0.057) & 0.830(0.053) & 0.781(0.052) & 0.830(0.051) & 0.792(0.042) & 0.814(0.057) \\
   ilpd-india &  583 &  10 & 0.697(0.045) & 0.702(0.050) & 0.685(0.056) & 0.681(0.072) & 0.666(0.050) & 0.649(0.039) & 0.669(0.045) & 0.649(0.042) \\
   mammograph &  961 &   6 & 0.830(0.024) & 0.831(0.022) & 0.836(0.023) & 0.833(0.031) & 0.802(0.038) & 0.837(0.028) & 0.827(0.028) & 0.823(0.035) \\
   molec-biol &  106 &  58 & 0.964(0.060) & 0.900(0.086) & 0.900(0.103) & 0.918(0.086) & 0.927(0.089) & 1.000(0.000) & 1.000(0.000) & 0.873(0.109) \\
      monks-1 &  556 &   7 & 0.988(0.016) & 0.629(0.046) & 0.825(0.051) & 0.845(0.042) & 0.893(0.040) & 0.995(0.008) & 0.986(0.013) & 0.973(0.022) \\
      monks-2 &  601 &   7 & 0.685(0.062) & 0.646(0.066) & 0.652(0.055) & 0.662(0.082) & 0.754(0.070) & 0.611(0.074) & 0.567(0.038) & 0.733(0.060) \\
      monks-3 &  554 &   7 & 0.977(0.014) & 0.714(0.067) & 0.963(0.028) & 0.955(0.018) & 0.889(0.047) & 0.988(0.011) & 0.954(0.033) & 0.968(0.024) \\
     mushroom & 8124 &  22 & 0.998(0.003) & 0.953(0.006) & 1.000(0.000) & 1.000(0.000) & 1.000(0.000) & 1.000(0.000) & 1.000(0.000) & 1.000(0.000) \\
       musk-1 &  476 & 167 & 1.000(0.000) & 0.992(0.014) & 0.988(0.019) & 0.973(0.013) & 0.904(0.038) & 0.996(0.012) & 0.996(0.012) & 0.990(0.014) \\
       musk-2 & 6598 & 167 & 0.998(0.004) & 1.000(0.000) & 1.000(0.000) & 0.998(0.002) & 0.977(0.006) & 1.000(0.000) & 1.000(0.000) & 1.000(0.000) \\
   oocytes\_me & 1022 &  42 & 0.846(0.024) & 0.784(0.022) & 0.838(0.020) & 0.779(0.022) & 0.722(0.035) & 0.780(0.025) & 0.755(0.040) & 0.841(0.023) \\
   oocytes\_tr &  912 &  26 & 0.837(0.015) & 0.774(0.035) & 0.822(0.039) & 0.823(0.034) & 0.728(0.062) & 0.817(0.035) & 0.778(0.036) & 0.830(0.027) \\
        ozone & 2536 &  73 & 0.973(0.011) & 0.972(0.007) & 0.972(0.009) & 0.972(0.010) & 0.971(0.011) & 0.970(0.009) & 0.973(0.008) & 0.970(0.008) \\
   parkinsons &  195 &  23 & 0.985(0.023) & 0.795(0.085) & 0.895(0.099) & 0.890(0.062) & 0.935(0.045) & 0.970(0.046) & 0.930(0.046) & 0.930(0.051) \\
          avg &      &     &        0.872 &        0.835 &        0.859 &        0.857 &        0.836 &        0.870 &        0.859 &        0.863 \\
       median &      &     &        0.898 &        0.832 &        0.864 &        0.850 &        0.863 &        0.900 &        0.892 &        0.873 \\
     avg rank &      &     &        5.569 &        4.224 &        4.741 &        4.500 &        2.983 &        5.224 &        4.207 &        4.552 \\
   \bottomrule
   \end{tabular}	
	\caption{Test accuracy for classification tasks on additional benchmark datasets, higher is better.}
	\label{table:class_acc}
\end{table}

\begin{table}[h!]
	\tiny
	\centering
	\begin{tabular}{lllllllllll}
   \toprule
      dataset &    N &   D &           OAK &              Linear &         SVGP &          SVM &          KNN &          GBM &           AdaBoost &          MLP \\
   \midrule
   acute-infl &  120 &   7 & -0.003(0.000) & -0.057(0.007) & -0.001(0.000) & -0.018(0.001) & -0.000(0.000) & -0.000(0.000) & -0.057(0.091) & -0.032(0.008) \\
   acute-neph &  120 &   7 & -0.003(0.003) & -0.032(0.009) & -0.001(0.000) & -0.019(0.001) & -0.000(0.000) & -0.000(0.000) & -0.085(0.256) & -0.017(0.004) \\
         bank & 4521 &  17 & -0.248(0.024) & -0.271(0.029) & -0.262(0.027) & -0.286(0.029) & -1.143(0.224) & -0.235(0.020) & -0.646(0.002) & -0.282(0.035) \\
        blood &  748 &   5 & -0.491(0.023) & -0.469(0.071) & -0.469(0.070) & -0.505(0.070) & -1.861(0.769) & -0.524(0.090) & -0.677(0.005) & -0.473(0.073) \\
   chess-krvk & 3196 &  37 & -0.078(0.041) & -0.056(0.009) & -0.051(0.011) & -0.020(0.008) & -0.232(0.088) & -0.010(0.015) & -0.010(0.007) & -0.020(0.018) \\
   congressio &  435 &  17 & -0.655(0.040) & -0.697(0.100) & -0.650(0.041) & -0.666(0.026) & -2.172(1.056) & -0.699(0.064) & -0.687(0.003) & -0.803(0.175) \\
   conn-bench &  208 &  61 & -0.040(0.026) & -0.090(0.070) & -0.084(0.092) & -0.077(0.040) & -0.208(0.073) & -0.000(0.000) & -0.000(0.000) & -0.189(0.117) \\
   credit-app &  690 &  16 & -0.280(0.126) & -0.365(0.083) & -0.369(0.085) & -0.377(0.070) & -1.182(0.552) & -0.111(0.074) & -0.583(0.008) & -0.363(0.080) \\
   cylinder-b &  512 &  36 & -0.475(0.055) & -0.533(0.059) & -0.532(0.033) & -0.463(0.045) & -0.888(0.375) & -0.392(0.049) & -0.654(0.007) & -0.530(0.149) \\
   echocardio &  131 &  11 & -0.358(0.170) & -0.394(0.185) & -0.376(0.157) & -0.423(0.167) & -1.107(1.274) & -0.444(0.273) & -0.584(0.024) & -0.385(0.202) \\
    fertility &  100 &  10 & -0.380(0.134) & -0.341(0.213) & -0.296(0.115) & -0.298(0.123) & -1.546(1.799) & -0.561(0.484) & -0.633(0.039) & -0.362(0.239) \\
   haberman-s &  306 &   4 & -0.532(0.099) & -0.531(0.093) & -0.530(0.106) & -0.540(0.094) & -1.468(1.395) & -0.570(0.157) & -0.679(0.009) & -0.540(0.114) \\
   heart-hung &  294 &  13 & -0.007(0.002) & -0.044(0.016) & -0.008(0.016) & -0.063(0.035) & -1.088(0.817) & -0.000(0.000) & -0.000(0.000) & -0.046(0.023) \\
    hepatitis &  155 &  20 & -0.414(0.105) & -0.389(0.100) & -0.346(0.065) & -0.352(0.077) & -1.306(0.798) & -0.531(0.159) & -0.570(0.046) & -0.362(0.130) \\
   hill-valle & 1212 & 101 & -0.694(0.001) & -0.650(0.013) & -0.693(0.000) & -0.694(0.001) & -1.498(0.294) & -0.708(0.027) & -0.693(0.004) & -0.675(0.013) \\
   horse-coli &  368 &  26 & -0.406(0.077) & -0.455(0.101) & -0.433(0.085) & -0.422(0.074) & -1.609(0.776) & -0.388(0.104) & -0.666(0.007) & -0.517(0.154) \\
   ilpd-india &  583 &  10 & -0.555(0.028) & -0.548(0.040) & -0.548(0.036) & -0.605(0.053) & -1.695(0.679) & -0.633(0.078) & -0.636(0.011) & -0.582(0.056) \\
   mammograph &  961 &   6 & -0.386(0.048) & -0.419(0.041) & -0.406(0.043) & -0.403(0.046) & -1.278(0.543) & -0.386(0.056) & -0.665(0.005) & -0.409(0.063) \\
   molec-biol &  106 &  58 & -0.149(0.172) & -0.211(0.138) & -0.203(0.136) & -0.196(0.171) & -0.283(0.100) & -0.000(0.000) & -0.000(0.000) & -0.363(0.186) \\
      monks-1 &  556 &   7 & -0.027(0.017) & -0.618(0.044) & -0.307(0.066) & -0.389(0.083) & -0.546(0.235) & -0.040(0.012) & -0.569(0.013) & -0.159(0.038) \\
      monks-2 &  601 &   7 & -0.549(0.063) & -0.648(0.044) & -0.638(0.048) & -0.589(0.056) & -0.676(0.398) & -0.653(0.085) & -0.679(0.007) & -0.505(0.056) \\
      monks-3 &  554 &   7 & -0.067(0.037) & -0.453(0.089) & -0.091(0.054) & -0.149(0.059) & -0.543(0.255) & -0.048(0.022) & -0.638(0.008) & -0.095(0.033) \\
     mushroom & 8124 &  22 & -0.009(0.021) & -0.135(0.009) & -0.002(0.001) & -0.000(0.000) & -0.000(0.000) & -0.003(0.000) & -0.483(0.006) & -0.001(0.000) \\
       musk-1 &  476 & 167 & -0.015(0.011) & -0.058(0.039) & -0.056(0.054) & -0.079(0.035) & -0.340(0.206) & -0.045(0.136) & -0.115(0.345) & -0.079(0.073) \\
       musk-2 & 6598 & 167 & -0.008(0.015) & -0.004(0.001) & -0.001(0.000) & -0.006(0.006) & -0.127(0.059) & -0.002(0.005) & -0.004(0.013) & -0.002(0.000) \\
   oocytes\_me & 1022 &  42 & -0.368(0.078) & -0.454(0.024) & -0.391(0.049) & -0.467(0.032) & -1.541(0.433) & -0.459(0.028) & -0.675(0.007) & -0.397(0.066) \\
   oocytes\_tr &  912 &  26 & -0.382(0.034) & -0.488(0.052) & -0.416(0.054) & -0.408(0.051) & -1.224(0.552) & -0.417(0.052) & -0.674(0.003) & -0.370(0.044) \\
        ozone & 2536 &  73 & -0.107(0.035) & -0.087(0.025) & -0.082(0.023) & -0.098(0.028) & -0.372(0.150) & -0.093(0.025) & -0.523(0.043) & -0.128(0.056) \\
   parkinsons &  195 &  23 & -0.065(0.050) & -0.320(0.094) & -0.207(0.102) & -0.253(0.090) & -0.150(0.043) & -0.256(0.399) & -0.439(0.047) & -0.196(0.041) \\
          avg &      &     &        -0.267 &        -0.338 &        -0.291 &        -0.306 &        -0.899 &        -0.283 &        -0.459 &        -0.306 \\
       median &      &     &        -0.280 &        -0.389 &        -0.307 &        -0.352 &        -1.088 &        -0.256 &        -0.584 &        -0.362 \\
     avg rank &      &     &         5.862 &         4.276 &         5.931 &         4.690 &         2.138 &         5.379 &         2.897 &         4.828 \\
   \bottomrule
   \end{tabular}
   	
	\caption{Test log likelihood for classification tasks on additional benchmark datasets, higher is better.}
	\label{table:class_ll}
\end{table}

\end{document}